\documentclass[]{style}
\usepackage[most]{tcolorbox}        
\usepackage{amsmath,amssymb}
\usepackage{lipsum}                 
\usepackage{xcolor}                 
\usepackage{fancyhdr}               
\usepackage{booktabs}
\usepackage[table]{xcolor}
\usepackage{url}
\usepackage{graphicx}   
\usepackage{subcaption} 
\usepackage{multirow}
\usepackage{algorithm,algorithmicx,algpseudocode} 
\usepackage{bm}
\definecolor{lightpurple}{RGB}{153, 102, 204}  
\definecolor{lilac}{RGB}{182, 133, 210}        
\usepackage{hyperref}
\hypersetup{
    colorlinks=true,      
    linkcolor=purple,        
    citecolor=lilac,       
    urlcolor=magenta      
}

\definecolor{abstractpurple}{HTML}{9C27B0} 

\renewcommand{\maketitle}{\mymaketitle}
\begingroup
\footnotetext{$*$ Equal contribution, $\dag$ Corresponding author}
\endgroup

\begin{document}

\title{RhymeFlow: Training-Free Acceleration for Video Generation with Asynchronous Denoising Flow Scheduling}
\author{Chensheng Dai$^{1,*}$, Shengjun Zhang$^{1,*}$, Yifan Li$^{1}$, Zhang Zhang$^{1}$, \\ Zheng Zhu$^{2}$, Yueqi Duan$^{1,\dag}$}
\affiliation[]{$^{1}$Tsinghua University, $^{2}$GigaAI}

\abstract{
Video generation models based on Diffusion Transformers (DiTs) have achieved remarkable performance in video synthesis, yet they suffer from high inference latency and computational costs due to the quadratic complexity of 3D attention.
Existing acceleration methods primarily reduce computational complexity within each individual denoising steps through techniques such as sparse attention and KV-caching.
However, they rigidly adhere to the inherent constraint of the standard diffusion pipeline: every frame in the target video sequence must be subjected to a complete, dense denoising process across all diffusion timesteps. 
We observe that due to the corresponding contents and motions among adjacent frames, when keyframes with critical semantic transitions are anchored, the intermediate states of others often follow more predictable trajectories, which indicates that such uniform, dense denoising process is inherently redundant for natural video data. 
To this end, we introduce \textbf{RhymeFlow}, a training-free framework that decouples the denoising trajectories of different frames.
Specifically, we first identify a sparse set of pivotal key frames that dominate the latent semantic evolution.
Then, only these keyframes undergo dense, step-by-step denoising to ensure structural integrity, while non-keyframes progressively skip denoising steps to minimize computational cost.
Since skipped intermediate states of non-keyframes break the temporal coherence in keyframe denoising steps, leading to visual degradation, we further introduce a latent trajectory projection module, which enables keyframes to interact with a complete and temporally consistent sequence representation.
Extensive experiments on current DiT-based video generation models demonstrate our method outperforms existing baselines with higher inference speed and better visual quality.
}

\checkdata[
\raisebox{-0.2em}{\includegraphics[width=0.025\linewidth]{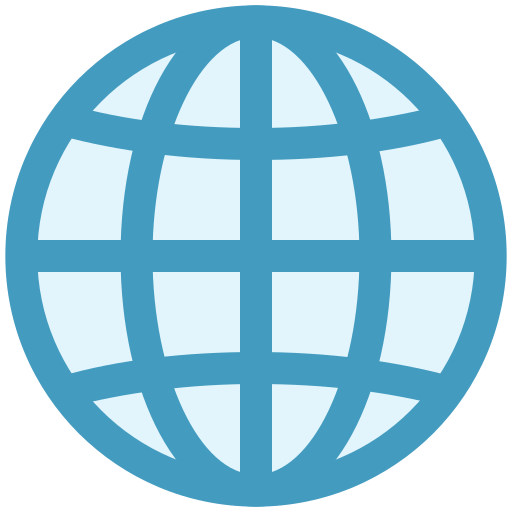}}~~Project Page]{\href{https://simon-dcs.github.io/Website-of-RhymeFlow/}{\texttt{https://simon-dcs.github.io/Website-of-RhymeFlow/}}
\\[-1.5ex]}

\checkdata[
\raisebox{-0.2em}{\includegraphics[width=0.025\linewidth]{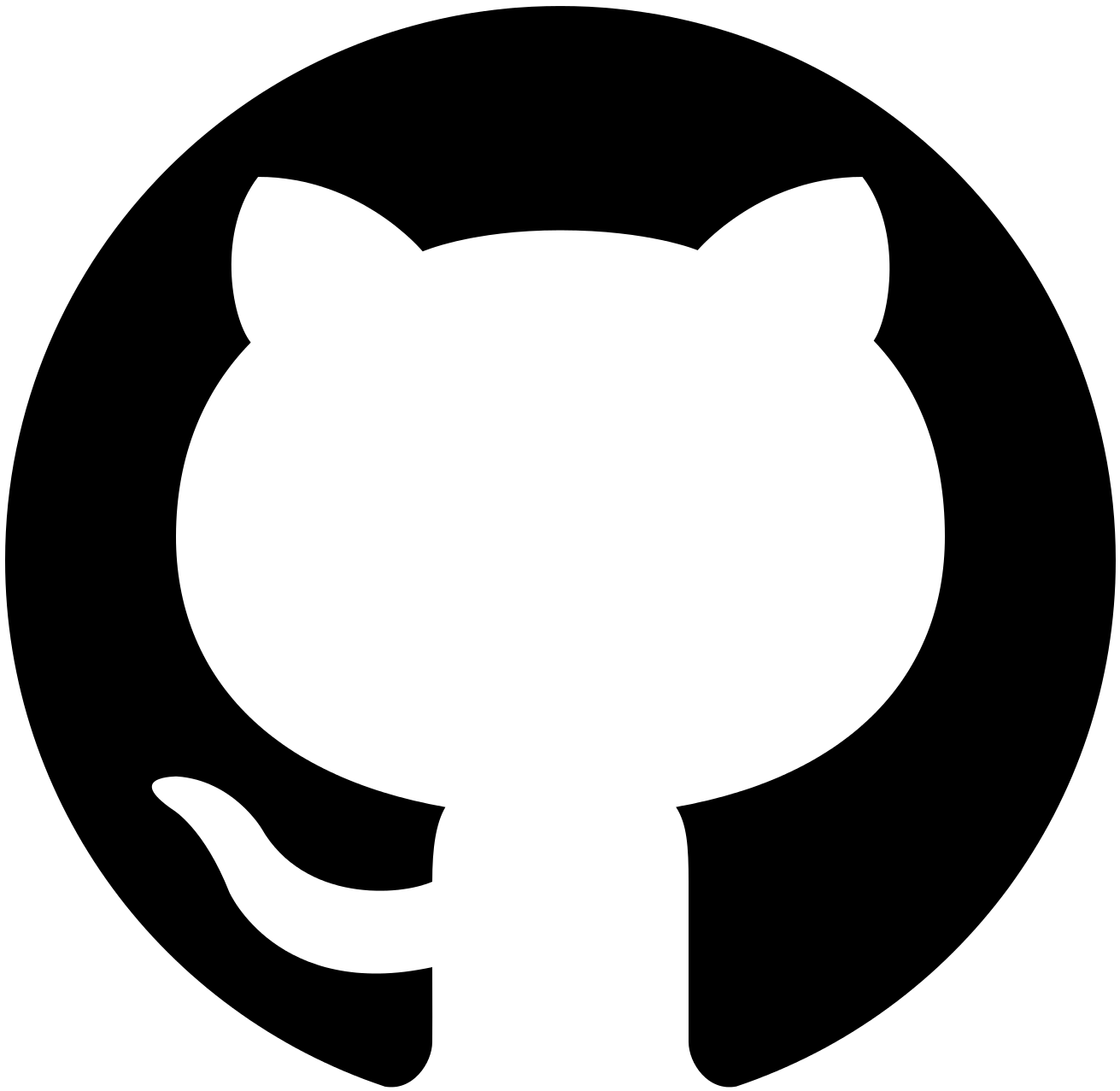}}~~GitHub Repo]{\href{https://github.com/Simon-Dcs/RhymeFlow}{\texttt{https://github.com/Simon-Dcs/RhymeFlow}}
\\[-1.5ex]}

\checkdata[
\raisebox{-0.2em}{\includegraphics[width=0.025\linewidth]{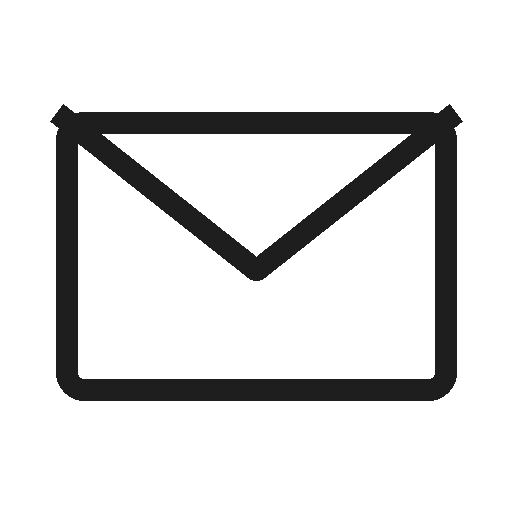}}~~Email]{\email{\{dcs23, zhangsj23\}@mails.tsinghua.edu.cn}
\\[-1.7ex]}

\maketitle

\section{Introduction}
\label{sec:intro}

\begin{figure*}[t]  
  \centering
  \includegraphics[width=\textwidth]{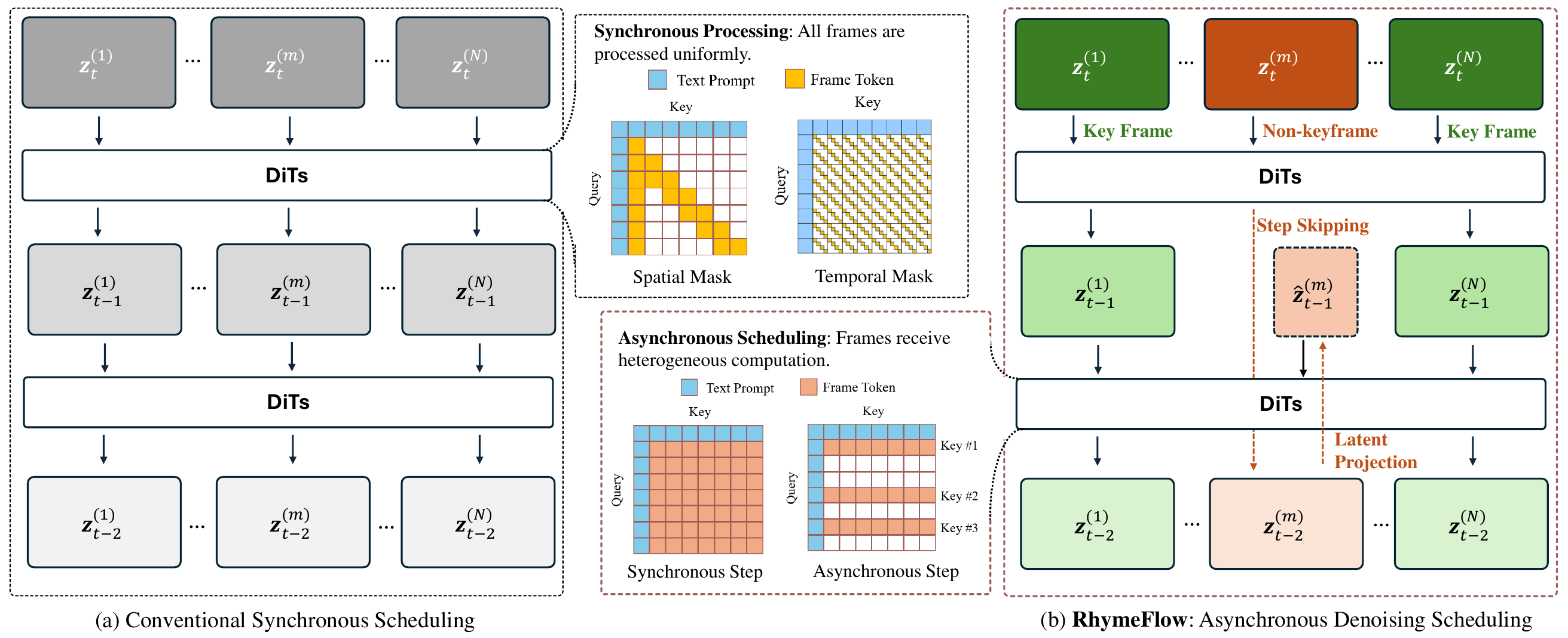}
  \caption{\textbf{Different sparse attention-based acceleration methods.} (a) Synchronous Scheduling: Conventional acceleration methods operate within a synchronous framework where all frames are jointly denoised at each timestep. These approaches achieve efficiency by exploiting intra-step sparsity through techniques such as spatial, temporal, or semantic attention masking.
  (b) Our Asynchronous Scheduling: Our work introduces an orthogonal acceleration dimension. Instead of processing all frames uniformly, we differentiate between keyframes and non-keyframes. Keyframes undergo a full, step-by-step denoising process to preserve high fidelity, while non-keyframes are updated asynchronously, effectively skipping computation at certain timesteps.}
  \label{fig2}
\end{figure*}


Video diffusion models~\cite{VideoDiffusion2022NIPS,xing2024survey,chen2024videocrafter2}, particularly those based on DiT architectures~\cite{DiT2023ICCV,ma2024latte,gupta2024photorealistic}, have achieved remarkable success in high-fidelity video synthesis. However, their practical deployment is hampered by a inherent bottleneck: the quadratic complexity of 3D spatiotemporal attention combined with dozens of denoising steps creates prohibitive inference costs~\cite{challenge2024,croitoru2023diffusion}. To mitigate this, training-free acceleration methods~\cite{freelong2024,freedom2023,trainingfree2024} have garnered significant attention due to their compatibility with pre-trained models, which broadly fall into three categories: KV-cache management~\cite{DeepCache2023arXiv,FastCache2024arXiv,Adaptivecache2025}, model compression via quantization or token pruning~\cite{SiTo2025AAAI,arquantization2024,tokenmerging2023}, and sparse attention mechanisms~\cite{SVG2025arXiv,SAP2025arXiv,sparseattntraining2025}. 
While these approaches effectively lower the complexity of every individual denoising step, every video frame is subjected to a dense, stepwise denoising procedure across the full set of diffusion timesteps.


In this work, we observe that such uniform computational allocation is inherently redundant.
Due to the overlapping visual contents and continuous motion trajectory among neighbor frames, video sequences natively exhibit strong spatiotemporal coherence.
Building upon this, we find that once a sparse set of pivotal keyframes—those capturing critical structural or semantic transitions—is anchored via step-by-step full attention, the intermediate states of adjacent non-keyframes often follow highly predictable trajectories in the latent space.
Driven by this predictability, allowing non-keyframes to appropriately skip certain denoising steps does not incur noticeable degradation in visual quality.
This observation indicates that applying a uniform, dense denoising schedule to every single frame is fundamentally unnecessary. 
Consequently, designing heterogeneous, frame-specific denoising schedules presents a promising, yet largely untapped, potential for video diffusion acceleration.

To this end, we introduce \textbf{RhymeFlow}, a training-free framework that establishes a new paradigm for video diffusion acceleration with \textbf{Asynchronous Denoising Flow Scheduling}. 
Our core insight is to decouple the denoising trajectories of individual frames and assign heterogeneous schedules accordingly. Specifically, we first identify a sparse set of pivotal keyframes by detecting semantic shifts in the latent space, ensuring that frames capturing critical transitions are prioritized for high-fidelity generation. Second, we assign heterogeneous denoising schedules: only these keyframes execute the complete, step-by-step denoising process to ensure structural integrity and high visual fidelity. In contrast, non-keyframes skip designated denoising steps to minimize computational costs. Third, recognizing that skipped intermediate states invariably break the temporal coherence required for 3D attention, we introduce a lightweight latent trajectory projection module. This mechanism analytically estimates the skipped intermediate latents of non-keyframes, which ensures keyframes can attend to a complete and temporally consistent sequence representation at every step, thereby preserving global consistency without incurring the penalty of full network evaluation.

\section{Related Works}
\label{sec:formatting}

\subsection{Video Diffusion Models}

Recent years have witnessed the remarkable success of diffusion models~\cite{DDPM2020NIPS,wijmans1995solution} in generating high-fidelity and diverse content. This paradigm has been successfully extended from images to the video domain, with Diffusion Transformers (DiTs)~\cite{DiT2023ICCV,ma2024latte,gupta2024photorealistic} emerging as the dominant architecture. State-of-the-art open-source models like Wan 2.1~\cite{Wan2025arXiv} , CogVideoX~\cite{CogVideo2022arXiv, CogVideoX2024arXiv}, and HunyuanVideo~\cite{Hunyuanvideo2024arXiv}, as well as closed-source systems like Sora~\cite{Sora2024} and Kling, have demonstrated unprecedented capabilities in generating temporally consistent and visually stunning videos from text or image prompts. These models typically adapt the 2D spatial attention mechanism to a more computationally intensive 3D spatiotemporal attention to capture both the appearance within frames and the motion across them~\cite{ViViT2021ICCV, Hunyuanvideo2024arXiv, CogVideoX2024arXiv}.

However, this success comes at a great cost. The inference process of these models is notoriously slow, often requiring hundreds of sequential passes through a massive transformer network. The 3D attention mechanism, in particular, exhibits quadratic complexity with respect to the number of tokens (pixels × frames), making the generation of long, high-resolution videos prohibitively expensive. This significant computational barrier motivates a strong need for efficient video generation techniques.





\subsection{Efficient Video Generation}

To combat the prohibitive inference costs, researchers have explored several avenues for acceleration. We categorize these efforts and position our work in relation to them.



\noindent\textbf{Decreasing the Denoising Steps.}  A primary strategy for accelerating diffusion models is to reduce the number of function evaluations (NFE) required during sampling. Early models based on stochastic differential equations (SDEs) required thousands of steps. The introduction of ordinary differential equation (ODE) solvers, such as Denoising Diffusion Implicit Models (DDIM)~\cite{DDIM2021ICLR}, and more advanced solvers like DPM-Solver~\cite{DPMSolver2022NIPS, DPMSloverv32023NIPS, DPMSlover++2025MIR}, significantly reduced the required steps. More recently, methods like Consistency Models~\cite{ConsistenctModels2023arXiv} and progressive distillation~\cite{ProgressiveDistillation2022ICLR} have been proposed to enable high-quality synthesis in just a few steps. However, these approaches have two major limitations in the context of large-scale video models. First, many of them, such as distillation and consistency training, require costly retraining or fine-tuning, which is impractical for models with billions of parameters. Second, they apply a uniform step-reduction strategy to all frames, treating them as equally important throughout the denoising process. 



\noindent\textbf{Diffusion Model Compression.}
An orthogonal line of research focuses on compressing the model itself to reduce the computational cost of each forward pass. This includes quantization techniques that reduce the bit-width of weights and activations, such as the W8A8 strategy in Q-Diffusion~\cite{QDiffusion2023ICCV}, or even more aggressive 4-bit schemes~\cite{SVDQuant2025ICCV}. Other approaches involve designing more efficient model architectures from the ground up or using more compact autoencoders.



\noindent\textbf{Training-free Acceleration via Caching.} 
A recent and related trend in training-free acceleration involves caching and reusing intermediate results. Methods like DeepCache~\cite{DeepCache2023arXiv} and FasterCache~\cite{FastCache2024arXiv} leverage the high feature similarity between adjacent denoising steps, avoiding redundant computation by reusing cached features (e.g., from the UNet's deeper layers or KV-cache). These methods are training-free and effectively reduce computation.



\noindent\textbf{Sparse Attention in LLMs and Video.}
The quadratic complexity of the attention mechanism has motivated extensive research into sparse attention, particularly in Large Language Models (LLMs). Methods like StreamingLLM~\cite{streamingllm}  and H2O~\cite{H2O2023NIPS}  identify that attention scores are often concentrated on a small subset of "heavy-hitter" or local tokens. This concept has been adapted for video diffusion. For instance, the proposed frameworks SVG~\cite{SVG2025arXiv} and SAP~\cite{SAP2025arXiv}, perform an in-depth analysis of attention patterns in Video DiTs, classifying heads into Spatial and Temporal types and applying structured sparse masks accordingly.
While highly effective, these methods focus on optimizing the computation within a single denoising step by reducing the number of token-to-token interactions. Instead of making the attention matrix sparse, RhymeFlow skips the entire forward pass for certain frames at certain steps, optimizing the denoising trajectory over time. This makes our asynchronous denoising flow scheduling paradigm orthogonal to intra-step sparsity. 

\section{Method}
\label{sec:method}

In this section, we preform a detailed introduction of \textbf{RhymeFlow}, a training-free acceleration framework for video diffusion models. 
Our approach re-envisions the denoising process by assigning different denoising schedule to different frames.
We first classify keyframes and non-keyframes based on the latent semantic evolution.
After the warm-up denoising stages, we propose a progressive skip strategy, where non-keyframes skip less steps at high noise levels and skip more steps at low noise levels.
To maintain the temporal coherence of natural video data, we further present a latent trajectory projection mechanism to enable keyframes to interact with non-keyframes at the denoising steps that they skip.

\subsection{Sequential Keyframe Selection}

To alleviate the prohibitive inference latency and excessive computational cost of prevailing video generation models while preserving the temporal continuity and semantic integrity of the generated sequence, we design a lightweight, content-aware sequential key frame selection scheme to identify representative frames that dominate the semantic transitions of the entire video.

Supposing that the video representation in the diffusion latent space consist of $N$ latent frames, we denote the noisy latent of the $i$-th frame at the current denoising timestep $t$ as $\mathbf{z}_t^{(i)}(1\leq t\leq N)$.
Since the raw noisy latents are heavily corrupted by noise and lack distinct structural information, we perform a single-step denoising prediction to estimate the fully denoised clean latent, denoted as $\hat{\mathbf{z}}_0^{(i)}$. These predicted clean latents provide a structurally clearer and more robust proxy for evaluating frame-wise correlations.

Based on the clean latent frames, we conduct the selection procedure in a strictly sequential manner along the temporal axis of the target video. 
Previous studies~\cite{ffgo,factorizing,consisti2v,kha_text2video} have comprehensively demonstrated that the initial frame of a video sequence acts as the fundamental semantic and visual anchor for subsequent frame synthesis, and plays an irreplaceable role in maintaining long-range temporal consistency and semantic fidelity throughout the generated video. 
In light of this, we initialize the set of selected key frames $\mathcal{K}=\{\hat{\mathbf{z}}_0^{(i)}\}$ with the initial frame of the target sequence. 
Subsequently, we traverse all subsequent candidate frames in chronological order. 
For each candidate latent frame $\hat{\mathbf{z}}_0^{(t)}$, we compute the pairwise similarity $\text{sim}(\hat{\mathbf{z}}_0^{(t)}, \hat{\mathbf{z}}_0^{(k)})$ between the candidate frame and the identified nearest preceding key frame $\hat{\mathbf{z}}_0^{(k)}\in \mathcal{K}$.
The update rule of the key frame set is written as:
\begin{equation}
\mathcal{K} \leftarrow
\begin{cases}
\mathcal{K} \cup \{\hat{\mathbf{z}}_0^{(t)}\}, & \text{if } \psi_\text{sim}(\hat{\mathbf{z}}_0^{(t)}, \hat{\mathbf{z}}_0^{(k)}) < \tau \\
\mathcal{K}, & \text{otherwise}
\end{cases}
\end{equation}
where $\psi_\text{sim}$ is defined as the cosine similarity function, and $\tau$ is a pre-defined threshold.

\begin{figure*}[t]  
  \centering
  \includegraphics[width=\textwidth]{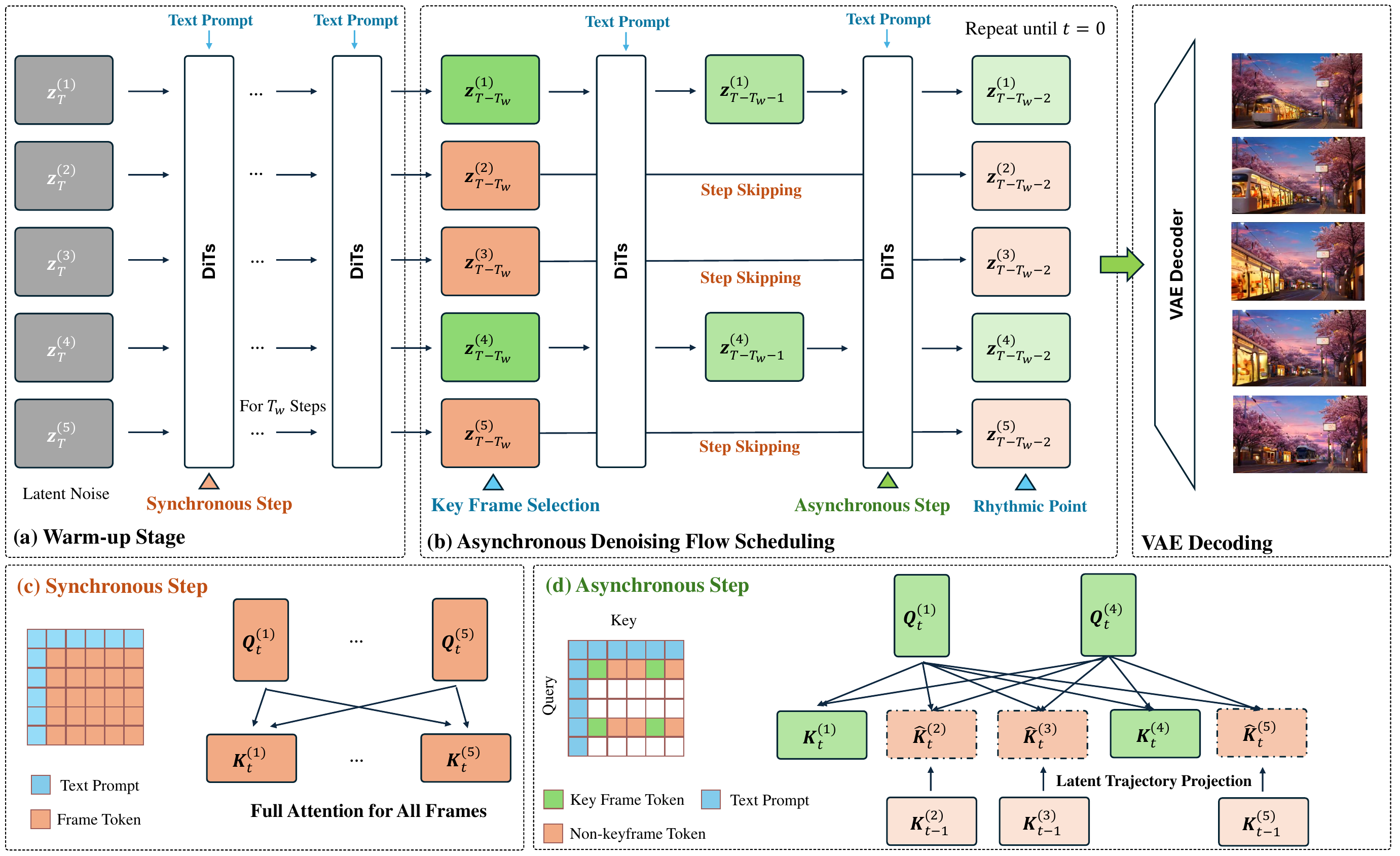}
  \caption{\textbf{Pipeline.}
  (a) \textbf{Warm-up Stage:} All frames are processed uniformly via synchronous updates for the initial $T_w$ steps. 
  (b) \textbf{Asynchronous Scheduling:} After warm-up, keyframes are selected. The pipeline transitions to an asynchronous schedule: keyframes are fully updated to preserve critical details, while non-keyframes are sparsely processed to accelerate generation. 
  (c) \textbf{Synchronous Step:} A standard full-attention mechanism is applied during warm-up and at "Rhythmic Points" to synchronize all latent frame representations.
  (d) \textbf{Asynchronous Step:} We introduce latent trajectory projection to efficiently estimate the skipped states of non-keyframes, providing the temporal context required for keyframes to attend to the full sequence.}
  \label{fig3}
  \vspace{-0.3cm}
\end{figure*}

\subsection{Asynchronous Denoising Flow Scheduling}

We design a heterogeneous denoising schedule where keyframes and non-keyframes are updated at different time-steps. 

\subsubsection{Initial Synchronous Warm-up}

The generation process commences with a synchronous warm-up stage that spans the initial $T_w$ denoising steps. This foundational phase is critical, as the early steps of the diffusion process are primarily responsible for establishing the global compositional structure, motion priors, and color palette of the nascent video. Bypassing full computation during this period can introduce severe, often irrecoverable, artifacts and quality degradation. Consequently, for the initial $T_w$ steps (i.e., from timestep $t=T$ down to $t=T-T_w$), all $N$ latent frames are updated simultaneously. The update for each latent frame $\mathbf{z}_t^i$ to $\mathbf{z}_{t-1}^i$ proceeds using the standard, unmodified denoising function with full attention, as illustrated in Figure \ref{fig3} (a).

\subsubsection{Progressive Asynchronous Scheduling}

The  reverse diffusion process exhibit strong non-uniformity along the timestep axis.
The early stages, characterized by high noise levels (large $t$), are dedicated to establishing the fundamental, low-frequency structure of the video, such as global composition and object forms. In this phase, the denoising trajectory is highly sensitive and less predictable, making these steps foundational to the final output quality. Conversely, the later stages (small $t$) primarily involve the refinement of high-frequency details, where the latent trajectory becomes significantly smoother and thus more predictable.

A static step-skipping stride cannot adapt to optimally navigate this evolving dynamic. 
An aggressive stride risks irrecoverable damage to the foundational structure in the early stages, while a conservative stride would forgo significant efficiency gains in the predictable later stages. 
To strike a deliberate balance between computational efficiency and generative fidelity, we introduce a Progressive Asynchronous Scheduling strategy. This approach adaptively increases the skipping stride for non-keyframes as the denoising process advances.

We implement this strategy with a simple, yet effective, piecewise schedule for the skip stride, $n_{\text{skip}}$, as a function of the current timestep $t$:
\begin{equation}
n_{\text{skip}}(t) = 
\begin{cases} 
  n_{\text{small}}, & \text{if } T_{\text{mid}} < t \le T - T_{w} \\
  n_{\text{large}}, & \text{if } t \le T_{\text{mid}}
\end{cases}
\label{eq:progressive_skip}
\end{equation}
where $T_{\text{mid}}$ is a mid-point hyperparameter (e.g., $T/2$), and $n_{\text{small}} < n_{\text{large}}$ are integer stride values (e.g., $n_{\text{small}}=2$, $n_{\text{large}}=3$). This  schedule intelligently allocates computational resources where they are most critical, thereby achieving a near-optimal trade-off between acceleration and the perceptual quality of the final generated video.

\subsubsection{Asynchronous Denoising Step}
\label{asyn}

Following the warm-up stage, RhymeFlow transitions into an asynchronous denoising schedule characterized by heterogeneous computational trajectories for different frame categories:
\begin{itemize}
    \item \textbf{Keyframes} are updated via a standard single-step denoising process, advancing their state from $\mathbf{z}_t^{(k)}$ to $\mathbf{z}_{t-1}^{(k)}$. This ensures they maintain fidelity.
    
    \item \textbf{Non-keyframes} are projected forward multiple steps in a single computation. Their state is advanced directly from $\mathbf{z}_t^{(k)}$ to $\mathbf{z}_{t-n_{\text{skip}}}^{(k)}$, thereby bypassing the intermediate denoising steps and accelerating the generation process.
\end{itemize}

Since keyframes and non-keyframes progress with staggered strides, they periodically coincide at specific timesteps, which we define as \textit{Rhythmic Points}. At these points, the framework executes a \textbf{full synchronous update}, wherein the latent representations of all frames engage in global 3D attention. This periodic re-synchronization provides a comprehensive contextual foundation, critical for maintaining global video coherence. Moreover, Rhythmic Points function as computational checkpoints where the trajectories of non-keyframes are recalibrated against the high-fidelity keyframe anchors. Such a mechanism effectively bounds the error accumulation inherent in prolonged step-skipping, ensuring that acceleration does not compromise temporal consistency or perceptual quality.

However, a fundamental challenge arises during the intermediate steps $\tau \in [t-n_{\text{skip}}+1, t-1]$, a keyframe $\mathbf{z}_{\tau}^{(k)} \in \mathcal{K}$ must be updated. This update requires attention context from all other frames at the target timestep $\tau$. The central challenge arises here: a non-keyframe $\mathbf{z}_{\tau}^{(j)}$ has been fast-forwarded to $t-n_{\text{skip}}$, meaning its latent states for any intermediate timestep $\tau$ are not explicitly computed and do not exist.

\subsubsection{Latent Trajectory Projection}
Inspired by the linear nature of ODE trajectories in Rectified Flow, we analytically project an approximation of the missing latent state. For a non-keyframe $j$ that was last updated at $t_{\text{start}}$ to obtain $\mathbf{z}_{t_{\text{end}}}^{(j)}$, we can project its latent state $\hat{\mathbf{z}}_{\tau}^{(j)}$ at the intermediate time $t$ via linear interpolation in the latent space:
\begin{equation}
\label{eq:projection}
\hat{\mathbf{z}}_{\tau}^{(j)} = (1-\alpha) \cdot \mathbf{z}_{t_{\text{start}}}^{(j)} + \alpha \cdot \mathbf{z}_{t_{\text{end}}}^{(j)}, \quad \alpha = \frac{t_{\text{start}} - \tau}{t_{\text{start}} - t_{\text{end}}}
\end{equation}
This projection is computationally trivial yet provides a robust estimate of the intermediate state. Following this latent projection, we can then compute the required Key and Value vectors, $\hat{\mathbf{K}}_{\tau}^{(j)}$ and $\hat{\mathbf{V}}_{\tau}^{(j)}$, on-the-fly.

With this mechanism, we define the logic for a keyframe update: it attends to the true states of other keyframes and the projected states of non-keyframes. A non-keyframe only performs its update at its designated, less frequent steps.

\subsubsection{KV-Cache Management.} Our asynchronous denoising schedule necessitates a specialized KV-cache mechanism fundamentally distinct from causal autoregressive approaches. We introduce a \textbf{per-layer rolling cache}  $\mathcal{C}_\ell^{(f)}$ for each DiT layer $\ell$ that stores the two most recent attention outputs for each non-keyframe $\hat{\mathbf{z}}$: $\mathbf{h}_{t_{1}}^{(\ell,\hat{\mathbf{z}})}$ and $\mathbf{h}_{t_{2}}^{(\ell,\hat{\mathbf{z}})}$ at timesteps $t_{\text{1}} > t_{\text{2}}$, along with their temporal indices. Critically, we cache the post-attention hidden states rather than input key-value pairs, as these representations have already integrated full-frame contextual information. When a keyframe requires context from a non-keyframe at a skipped timestep $\tau \in (t_{\text{1}}, t_{\text{2}})$, we employ latent trajectory projection. These interpolated representations serve as transient key-value providers for the current attention operation and are immediately discarded thereafter, eliminating persistent storage of intermediate states.

\section{Experiments}
\label{exp}

\begin{table}[t] 
\centering
\caption{\textbf{Quantitative evaluation of the efficiency-fidelity trade-off on Wan 2.1.} We investigate the impact of warm-up timesteps ($T_w$) and the number of keyframes ($M$). Our optimal default configuration is highlighted in bold.}
\label{table1}

\begin{tabular}{cc ccc cc cc} 
\toprule
\multicolumn{2}{c}{Method} & \multicolumn{3}{c}{Similarity Metrics} & \multicolumn{2}{c}{Video Quality} & \multicolumn{2}{c}{Efficiency} \\
\cmidrule(lr){1-2} \cmidrule(lr){3-5} \cmidrule(lr){6-7} \cmidrule(lr){8-9}
$T_w$ & $M$ & PSNR $\uparrow$ & SSIM $\uparrow$ & LPIPS $\downarrow$ & SubCon. $\uparrow$ & ImgQual. $\uparrow$ & Lat.(s) $\downarrow$ & Speedup $\uparrow$\\
\midrule
\multicolumn{2}{c}{Original} & - & - & - & 0.9102  &  0.6946 & 993.5 & - \\
\midrule
6 & 3  & 20.998 & 0.615 & 0.206 & 0.8460  & 0.6099  &  607.2 &  1.64$\times$ \\
6 & 4  & 22.588 & 0.617 & 0.196 &  0.8478 &  0.6362 & 642.0 & 1.55$\times$\\
6 & 5 & 25.669 & 0.657 & 0.184 &  0.8536 &  0.6883 & 682.0 & 1.46$\times$\\
\midrule
8 & 3  & 23.742 & 0.721 & 0.182 &  0.8601 & 0.6344  &  621.6 & 1.60$\times$  \\
\textbf{8} & \textbf{4}  & \textbf{26.291} & \textbf{0.783} & \textbf{0.168} &  \textbf{0.8831} & \textbf{0.6706}  & \textbf{650.4}  &  \textbf{1.53$\times$} \\
8 & 5 & 27.707 & 0.812 & 0.162 &  0.8896 &  0.6902 & 712.8 &  1.39$\times$\\
\midrule
10 & 3  & 24.006 & 0.742 & 0.169 &  0.8818 & 0.6714 & 630.8 & 1.57$\times$ \\
10 & 4  & 26.636 & 0.817 & 0.164 &  0.8838  &  0.6819 & 670.3 & 1.48$\times$ \\
10 & 5 & 28.061 & 0.816 & 0.161 &  0.8911 &  0.6919 & 738.1 &  1.35$\times$\\
\bottomrule
\end{tabular}

\end{table}

\subsection{Experimental Settings}
\label{settings}

\begin{figure*}[t]
  \centering
  \includegraphics[width=\textwidth]{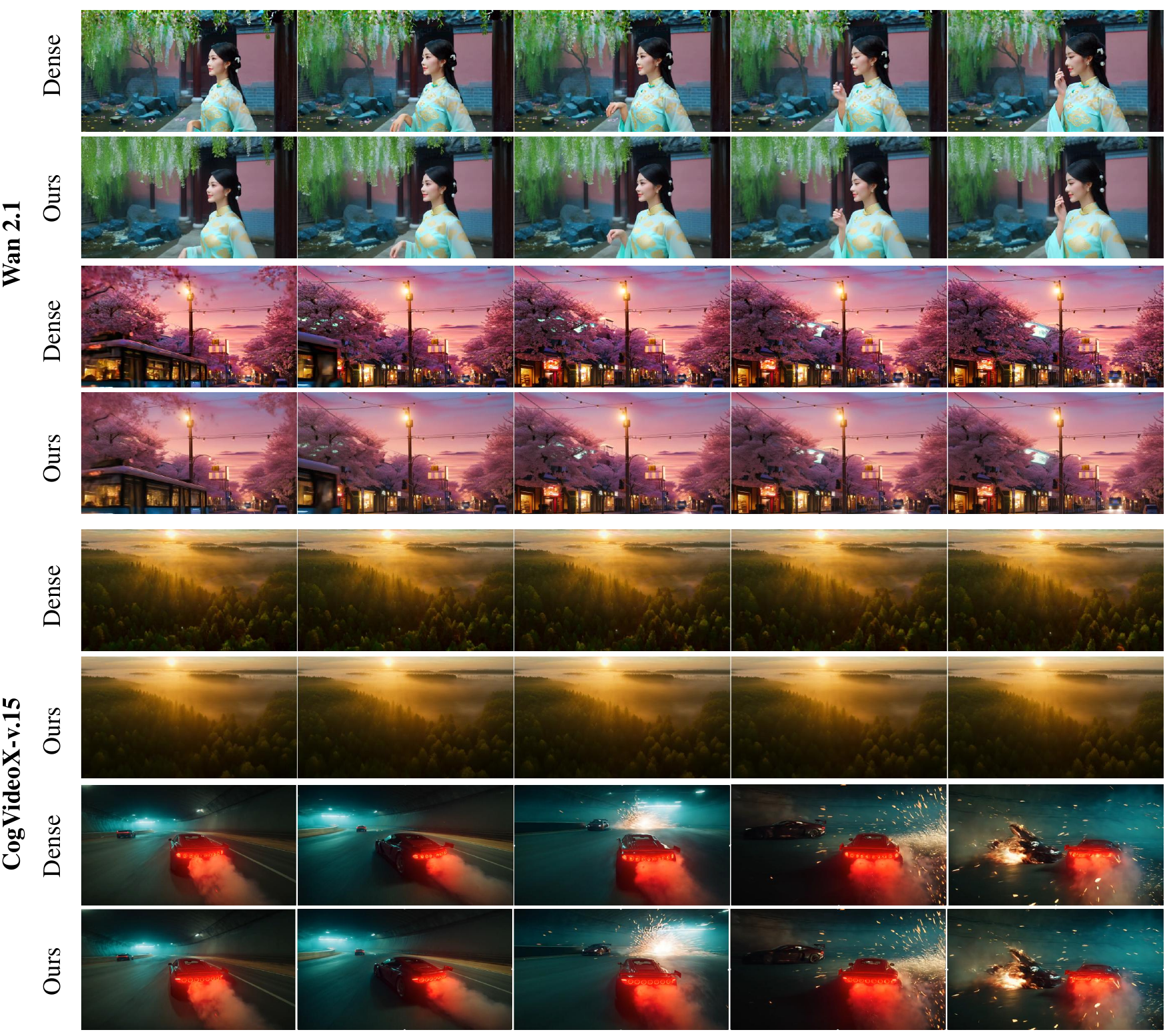}
  \caption{\textbf{Qualitative visual comparisons.} Frame sequences generated by the standard dense attention baseline versus our RhymeFlow. The top two examples are produced by Wan 2.1, while the bottom two are from CogVideoX-v1.5.}
  \label{fig4}
\end{figure*}

\begin{table}[t]
\centering
\caption{\textbf{Quality and efficiency benchmarking on Wan 2.1 and CogVideoX-v1.5.}}
\label{table2}
\resizebox{\textwidth}{!}{
\begin{tabular}{l|ccccc|cc}
\toprule
Method & \multicolumn{5}{c|}{Quality} & \multicolumn{2}{c}{Efficiency} \\
\cmidrule{2-8}
 & PSNR $\uparrow$ & SSIM $\uparrow$ & LPIPS $\downarrow$ & SubCon. $\uparrow$ & ImgQual. $\uparrow$ & Lat.(s) $\downarrow$ & Speedup $\uparrow$ \\
\midrule
Wan 2.1  & - & - & - & 0.9102 & 0.6946 & 993.5 & - \\
\midrule
SpargeAttn \cite{SpargeAttn2025ICML} & 20.399 & 0.613 & 0.393 & 0.8632 & 0.7118 & 719.7 & 1.38$\times$ \\
SVG \cite{SVG2025arXiv} & 22.419 & 0.694 & 0.290 & 0.8758 & 0.6913 & 708.0 & 1.40$\times$ \\
SAP \cite{SAP2025arXiv} & 24.454 & 0.730 & 0.223 & 0.8789 & 0.6837 & 608.5 & 1.63$\times$ \\
\textbf{Ours} & \textbf{26.291} & \textbf{0.783} & \textbf{0.168} & \textbf{0.8831} & \textbf{0.6706} & \textbf{650.4} & \textbf{1.53$\times$} \\
\textbf{Ours + SAP} & \textbf{24.586} & \textbf{0.737} & \textbf{0.221} & \textbf{0.8792} & \textbf{0.6806} & \textbf{596.8} & \textbf{1.66$\times$} \\
\midrule
CogVideoX-v1.5  & - & - & - & 0.987  & 0.624  &  625.0 & - \\
\midrule
MInference ~\cite{jiang2024minference} & 22.490 & 0.743 & 0.264 & 0.874 & 0.589 & 422.3 & 1.48$\times$ \\
PAB ~\cite{zhao2024real} & 23.230 & 0.782 & 0.145 & 0.978 & 0.573 & 443.3 & 1.41$\times$ \\
SVG \cite{SVG2025arXiv} & 24.130 & 0.811 & 0.171 & 0.982 & 0.597 & 385.8 & 1.62$\times$ \\
\textbf{Ours} & \textbf{26.890} & \textbf{0.852} & \textbf{0.142} & \textbf{0.986} & \textbf{0.623} & \textbf{351.1} & \textbf{1.78$\times$} \\
\textbf{Ours + SAP} & \textbf{25.574}  &  \textbf{0.821} &  \textbf{0.157} & \textbf{0.972} & \textbf{0.609} & \textbf{323.8} & \textbf{1.93$\times$} \\
\bottomrule
\end{tabular}
}
\end{table}

\textbf{Models.}
To validate the effectiveness of our proposed framework, we integrate it into two prominent open-source text-to-video (T2V) generation models: Wan2.1-T2V-v1.3B-Diffusers~\cite{Wan2025arXiv} and CogVideoX-v1.5-T2V ~\cite{CogVideoX2024arXiv}. Both models are configured to produce videos of 81 frames at 720p resolution. The architectural specifics of these models are noteworthy: in its latent space, Wan 2.1 operates on 21 latent frames, with each frame represented by 3600 tokens post 3D-VAE processing. In contrast, CogVideoX-v1.5 utilizes a 3D full attention mechanism over 11 frames, where each frame corresponds to 4080 tokens after its 3D-VAE. 

\textbf{Metrics.}
We evaluate our method across two primary dimensions: generation fidelity and computational efficiency. To quantify fidelity preservation, we compare the accelerated outputs against the unaccelerated baselines using standard pixel-level and perceptual metrics: Peak Signal-to-Noise Ratio (PSNR), Structural Similarity (SSIM), and Learned Perceptual Image Patch Similarity (LPIPS). Additionally, we assess the visual quality of the generated videos using VBench~\cite{VBench2024CVPR}, focusing on critical aspects such as subject consistency and imaging quality. Finally, to demonstrate computational efficiency, we report the average inference latency of diffusion process alongside the corresponding speed-up ratio.

\textbf{Baselines.}
To contextualize the performance of RhymeFlow, we conduct a comparative analysis against several state-of-the-art (SOTA) training-free acceleration methods. Our baselines include SpargeAttn ~\cite{SpargeAttn2025ICML}, MInference ~\cite{jiang2024minference}, Pyramid Attention Broadcast (PAB) ~\cite{zhao2024real}, Sparse VideoGen (SVG) ~\cite{SVG2025arXiv}, and Semantic-Aware Permutation (SAP) ~\cite{SAP2025arXiv}. For a fair comparison, we adopt the official hyperparameter configurations provided by the authors for their respective text-to-video generation tasks. All experiments are executed on a single NVIDIA A800 GPU to ensure a consistent hardware environment.

\textbf{Implementation Details.}
Our text-to-video generation experiments utilize prompts sourced from the Penguin Benchmark, which have been refined through the prompt optimization process detailed in VBench~\cite{VBench2024CVPR}. 
The core hyperparameters are configured as follows: the duration of the warm-up stage, $T_w$, is set to 8 steps, and the total number of keyframes, $M$, is 4. For the progressive asynchronous scheduling, the update strides are defined as $n_{\text{small}} = 2$ and $n_{\text{large}} = 3$.

\begin{table}[t] 
\centering
\caption{\textbf{Quality and efficeincy benchmarking on HunyuanVideo.}}
\label{retable1}

\begin{tabular}{l|ccccc}
\toprule
Method & PSNR $\uparrow$ & SSIM $\uparrow$ & LPIPS $\downarrow$ & Latency (s) $\downarrow$ & Speedup $\uparrow$\\
\midrule
SVG~\cite{SVG2025arXiv}   & 21.17 & 0.684 & 0.392 &   3459  &   1.92 $\times$ \\
SAP~\cite{SAP2025arXiv}   &  24.64 &  0.904 &  0.068 &  2634 &  2.52 $\times$\\
EasyCache~\cite{zhou2025easycache} &  23.51  &  0.861   &  0.119   &   2850  &   2.33 $\times$ \\
DiCache~\cite{bu2025dicache} &  23.54  &  0.860   &  0.114   &  2814   &   2.36 $\times$ \\
VGDFR~\cite{VGDFR2025arXiv} &  19.49  &  0.779   &  0.199   &   3019  &   2.20 $\times$ \\
\textbf{Ours}  & \textbf{26.34} & \textbf{0.918} & \textbf{0.060} &  2939  & 2.26  $\times$\\
\textbf{Ours+SAP}  &  25.01 & 0.910 & 0.068 & \textbf{2555} & \textbf{2.60 $\times$} \\
\bottomrule
\end{tabular}

\end{table}

\begin{table}[t]
\centering
\caption{\textbf{Ablation on key frame selection strategy.} }
\label{tab:ablation_keyframe}
\begin{tabular}{l ccccc} 
\toprule 
Setup & PSNR $\uparrow$ & SSIM $\uparrow$ & LPIPS $\downarrow$ & Latency $\downarrow$ & Speedup $\uparrow$ \\ 
\midrule
Original & - & - & - & 993.5 & - \\
\midrule
Random  & 20.630 & 0.525 & 0.383 & 650.0  &  1.53$\times$ \\
First   & 19.220 & 0.515 & 0.402 & 651.0  &  1.53$\times$ \\
Uniform & 24.293 & 0.643 & 0.183 & 649.0  &  1.53$\times$ \\
Ours    & 26.291 & 0.783 & 0.168 & 650.4  &  1.53$\times$ \\
\bottomrule
\end{tabular}
\end{table}

\begin{table}[t]
\centering
\caption{\textbf{Ablation study on Architectural Components.} }
\label{tab:ablation_component}
\begin{tabular}{l ccccc}
\toprule
Method & PSNR $\uparrow$ &SSIM $\uparrow$ & LPIPS $\downarrow$ & Latency $\downarrow$ & Speedup $\uparrow$ \\ 
\midrule
Original & - & - & - & 993.5 & - \\
\midrule
w/o Progressive & 25.399 & 0.753 & 0.172 & 708.0 & 1.40$\times$ \\
w/o Projection  & 20.630 & 0.525 & 0.383 & 622.2 & 1.60$\times$ \\
Ours            & 26.291 & 0.783 & 0.168 & 650.4  &  1.53$\times$ \\
\bottomrule
\end{tabular}
\end{table}

\begin{table}[t] 
\centering
\caption{\textbf{Ablation study on KV-Cache management.}}
\label{retable2}
{
\begin{tabular}{lcccc}
\toprule
Method & Latency (s) $\downarrow$ & Speedup $\uparrow$ & Average Memory (GB) $\downarrow$ & Peak Memory (GB) $\downarrow$ \\
\midrule
Original & 993.5 & 1.00$\times$ & 24.4 & 44.3 \\
\midrule
w/o KV-Cache & 653.6 & 1.52$\times$ & 27.1 & 49.9 \\
Ours & 650.4 & 1.53$\times$ & 21.5 & 42.6 \\
\bottomrule
\end{tabular}
}
\end{table}

\begin{table}[!t]
\centering
\caption{\textbf{Latent-state analysis against dense sampler on Wan 2.1.}}
\label{tab: smapler ablation}
\begin{tabular}{lccccc}
\toprule
Variant & PSNR\,$\uparrow$ 
& $\mathcal{E}_\text{keyframe}\,\downarrow$ 
& $\mathcal{E}_\text{non-keyframe}\,\downarrow$ 
& $\mathcal{E}_\text{projected-nonkeyframe}\,\downarrow$ 
& Speedup\,$\uparrow$ \\
\midrule
Ours (full)                        & \textbf{27.70} & \textbf{0.0019} & \textbf{0.0022} & \textbf{0.0018} & 1.44$\times$ \\
\;\;w/o latent projection          & 24.44          & 0.0190         & 0.0029          & 0.0240         & 1.55$\times$ \\
\;\;w/ keyframe-only attention     & 24.37          & 0.0043          & 0.0045          & 0.0490         & \textbf{1.69}$\times$ \\
\bottomrule
\end{tabular}
\end{table}

\subsection{Trade-off Analysis of Scheduling Parameters}
\label{sec:quality_tradeoff}

To evaluate our asynchronous scheduling, we analyze the efficiency-fidelity trade-off by varying the warm-up duration ($T_w$) and keyframe budget ($M$). As shown in Table \ref{table1}, both parameters predictably trade acceleration for generation quality. 

Specifically, extending $T_w$ strengthens the global structure, steadily improving Subject Consistency (SubCon.) and Image Quality (ImgQual.). Concurrently, increasing $M$ provides more frequent high-fidelity temporal anchors, which effectively reduces error accumulation; for instance, at a fixed $T_w=8$, raising $M$ from $3$ to $5$ boosts PSNR from 23.742 to 27.707 and SSIM from 0.721 to 0.812, though the inference speedup drops from $1.60\times$ to $1.39\times$. 

Empirically, we identify $T_w=8$ and $M=4$ as the optimal configuration. This setting preserves strong visual and temporal fidelity (ImgQual. 0.6706, SubCon. 0.8831) comparable to the original dense model, while delivering a substantial $1.53\times$ inference speedup (650.4s vs. 993.5s). Consequently, we adopt this configuration as our default for all subsequent benchmark comparisons.

\subsection{Comparison with State-of-the-Art Methods}
\label{efficiency}

We benchmark RhymeFlow against several state-of-the-art training-free video acceleration methods across two fundamentally different architectures: Wan 2.1 and CogVideoX-v1.5. To ensure a fair and rigorous evaluation, we select baseline methods based on their official architectural compatibility. For instance, SAP \cite{SAP2025arXiv} is evaluated strictly on Wan 2.1, while MInference \cite{jiang2024minference} and PAB \cite{zhao2024real} are evaluated on CogVideoX-v1.5, mitigating any performance degradation caused by unverified cross-architecture adaptation.

\subsubsection{Quantitative Evaluation}
As detailed in Table \ref{table2}, RhymeFlow consistently establishes a superior trade-off between generation quality and efficiency. On Wan 2.1~\cite{Wan2025arXiv}, RhymeFlow achieves a high PSNR of 26.291 and an SSIM of 0.783 with a $1.53\times$ speedup, significantly outperforming intra-step methods like SpargeAttn~\cite{SpargeAttn2025ICML} and standalone SVG~\cite{SVG2025arXiv} in visual fidelity. This advantage is equally evident on CogVideoX-v1.5~\cite{CogVideoX2024arXiv}, where RhymeFlow attains a leading $1.78\times$ speedup while preserving a remarkable SubConsistency of 0.986.
We further benchmark against EasyCache~\cite{zhou2025easycache}, DiCache~\cite{zhou2025easycache}, and VGDFR~\cite{VGDFR2025arXiv} on high-dynamic HunyuanVideo~\cite{Hunyuanvideo2024arXiv} clips in Table~\ref{retable1}. 
Our base method achieves the best visual quality across all metrics, outperforming cache-based methods, such as EasyCache~\cite{zhou2025easycache} and DiCache~\cite{bu2025dicache}. When combined with SAP (Ours+SAP), our approach achieves the lowest latency and highest speedup.

\subsubsection{Orthogonality and Synergistic Combination} 
As discussed in Section \ref{discussonorg}, our asynchronous frame-scheduling operates inter-step, making it inherently orthogonal to token-level (intra-step) sparse attention methods. To demonstrate this, we evaluate a hybrid approach, denoted as \textit{Ours + SAP}. Notably, standard SAP imposes an aggressive sparsity ratio of 0.3. Because RhymeFlow already significantly reduces the computational burden by skipping frame-level updates, applying extreme token-level sparsity concurrently would lead to information bottleneck. Therefore, we deliberately relax the SAP~\cite{SAP2025arXiv} sparsity ratio to 0.5 in the hybrid setting. This synergistic parameter scheduling allows \textit{Ours + SAP} to not only achieve the fastest generation speeds ($1.66\times$ on Wan 2.1~\cite{Wan2025arXiv} and $1.93\times$ on CogVideoX-v1.5~\cite{CogVideoX2024arXiv}) but also yield better generation quality than the standalone SAP~\cite{SAP2025arXiv} baseline.

\subsubsection{Qualitative Results}
The quantitative superiority translates directly into visual fidelity. As illustrated in Figure \ref{fig4}, we present qualitative comparisons between the standard dense baseline and RhymeFlow across a diverse array of challenging scenarios. From top to bottom, these include fine-grained human portraits, structurally complex urban cityscapes, nuanced natural environments, and high-dynamic motion scenes. Across all settings, RhymeFlow perfectly maintains the intricate textures, lighting consistency, and complex motion dynamics of the dense baseline without introducing the flickering or blurring artifacts commonly associated with aggressive acceleration strategies.

\subsection{Ablation Study}
\label{ablation}

To thoroughly validate the design choices of RhymeFlow, we conduct extensive ablation studies, analyzing keyframe selection, core framework components, and memory management on Wan 2.1~\cite{Wan2025arXiv}.

\subsubsection{Ablation on Key Frame Selection Strategy} 
We first evaluate our content-aware keyframe selection mechanism against three heuristic alternatives with a fixed budget of $M=7$: \textit{Random}, \textit{First} (initial $M$ frames), and \textit{Uniform} (evenly spaced intervals). 

As presented in Table \ref{tab:ablation_keyframe}, naive placement strategies like \textit{Random} and \textit{First} lead to severe visual degradation (PSNR drops below 21). The \textit{Uniform} strategy serves as a much stronger baseline by providing a basic temporal coverage, achieving a PSNR of 24.293. However, because it blindly assigns keyframes without considering the actual semantic shifts within the video, it struggles to maintain high-frequency details across dynamic scenes. By dynamically identifying frames with substantial latent changes, our method significantly outperforms the \textit{Uniform} baseline, boosting SSIM from 0.643 to 0.783 and lowering LPIPS from 0.183 to 0.168. This highlights that an optimal strategy must be context-aware, placing high-fidelity anchors exactly where motion or semantic transitions occur.

\begin{figure}[!t]
    \centering
    \includegraphics[width=\linewidth]{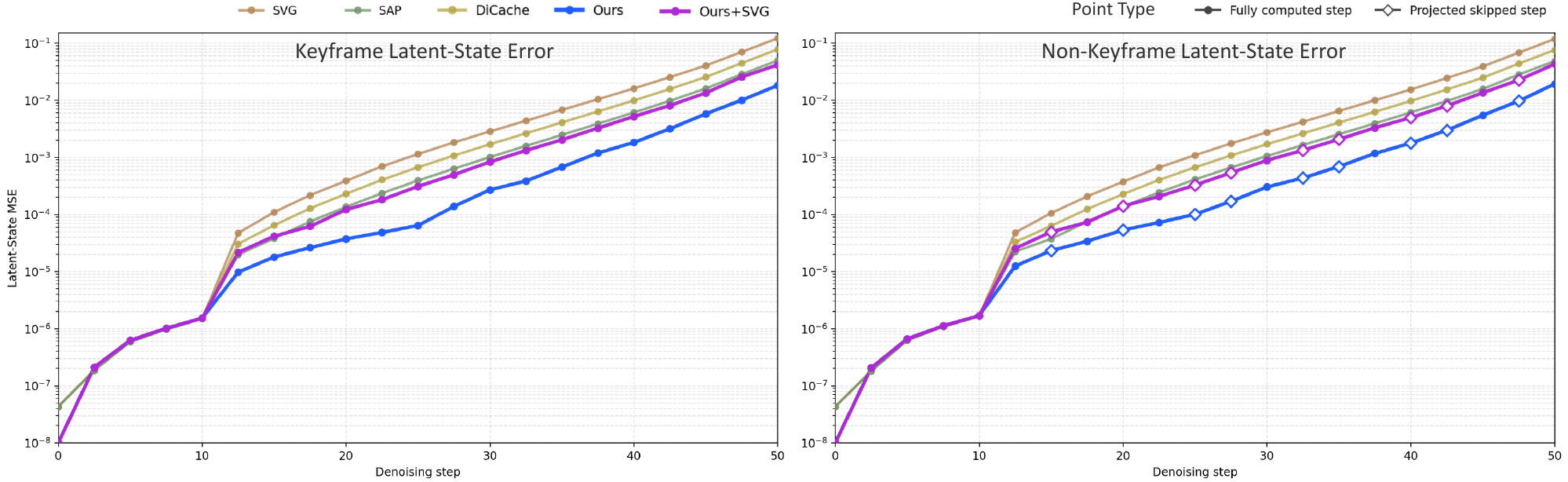}
    \caption{\textbf{The error analysis of the latent projection on high-dynamic videos.}}
    \label{fig:error analysis}
\end{figure}

\subsubsection{Ablation on Core Architectural Components} 
Table \ref{tab:ablation_component} isolates the contributions of our core algorithmic designs. Replacing progressive scheduling with a fixed-step skip (\textit{w/o Progressive}) diminishes both acceleration ($1.53\times \rightarrow 1.40\times$) and visual fidelity (PSNR 26.291 $\rightarrow$ 25.399). This validates our dynamic stride, which optimally preserves dense updates in noise-sensitive early stages while aggressively skipping predictable later stages. Conversely, omitting trajectory projection (\textit{w/o Projection}) forces keyframes to attend only to the sparse synchronization group. Although this yields a marginal speedup ($1.60\times$), depriving the model of dense temporal context triggers a quality collapse (PSNR plummets to 20.630). Thus, this computationally lightweight projection is indispensable for stabilizing asynchronous diffusion.

\subsubsection{Efficacy of KV-Cache Management} 

Table \ref{retable2} demonstrates that our KV-Cache management is essential for suppressing memory overhead. While both configurations (\textit{w/o KV-Cache} and \textit{Ours}) achieve similar speedups ($1.52\times$ vs. $1.53\times$) natively driven by asynchronous step-skipping, the naive implementation (\textit{w/o KV-Cache}) must retain historical intermediate states to perform future latent projections. This persistent storage severely bloats the peak VRAM to 49.9 GB, exceeding even the original dense model (44.3 GB). By introducing a per-layer rolling cache that instantly discards transient interpolated features, RhymeFlow efficiently bounds the peak memory to 42.6 GB. This deliberate design ensures accessible GPU deployment without sacrificing acceleration.

\subsubsection{Latent Projection Analysis}
The near-straight per-frame trajectories induced by flow matching keep the linear-projection error small throughout denoising. Figure~\ref{fig:error analysis} plots per-step latent MSE, and Table~\ref{tab: smapler ablation} shows that removing projection or non-keyframe context increases mean latent error and reduces PSNR.

\subsection{User Study}
We conducted a 82-participant user study for human preference evaluation.
The user study results are summarised in Table~\ref{retab:user_study}. 
For each comparison, $p$-values were computed using binomial tests after excluding ties. 
For the comparisons against SVG and SAP, we used one-sided tests ($H_1$: Ours $>$ Opponent); 
for the comparison against Dense, a two-sided test ($H_0$: no difference) was applied.
Our method outperformed SVG in all three metrics, with statistically significant differences. 
For visual quality, 51.2\% of participants preferred ours versus 30.5\% for SVG. 
Temporal coherence favoured ours by 53.7\% to 20.7\%. 
Our method achieved even larger margins over SAP, where all differences are highly significant.
Against Dense, the results are not statistically significant. 
The two-sided tests indicate no evidence of a difference between our method and Dense across all metrics.

\begin{table}[t]
\centering
\caption{\textbf{Double-blind user study results (\%).}}
\label{retab:user_study}
\setlength{\tabcolsep}{3.5pt}
\begin{tabular}{@{} l cccc cccc cccc @{}}
\toprule
\multirow{2}{*}{\textbf{Metric}} & \multicolumn{4}{c}{\textbf{Ours v.s.\ SVG}} & \multicolumn{4}{c}{\textbf{Ours v.s.\ SAP}} & \multicolumn{4}{c}{\textbf{Ours v.s.\ Dense}} \\
\cmidrule(lr){2-5} \cmidrule(lr){6-9} \cmidrule(lr){10-13}
& Ours & Tie & Opp. & $p$-Value & Ours & Tie & Opp. & $p$-Value & Ours & Tie & Opp. & $p$-Value \\
\midrule
Visual Quality     & \textbf{51.2} & 18.3 & 30.5 & .025 & \textbf{74.4} & 14.6 & 11.0 & $<$\!.001 & 18.3 & \textbf{53.7} & 28.0 & .256 \\
Temporal Coherence & \textbf{53.7} & 25.6 & 20.7 & $<$\!.001 & \textbf{78.0} & 12.2 & 9.8 & $<$\!.001 & 18.3 & \textbf{58.5} & 23.2 & .608 \\
Perceptual Preference   & \textbf{51.2} & 23.2 & 25.6 & .006 & \textbf{79.3} & 12.2 & 8.5 & $<$\!.001 & 15.9 & \textbf{56.1} & 28.0 & .133 \\
\bottomrule
\end{tabular}
\end{table}
\section{Conclusion}



In this paper, we present \textbf{RhymeFlow}, a training-free acceleration framework that revisits the computational paradigm of video diffusion models. Instead of synchronously updating all frames at every denoising step, RhymeFlow introduces a content-aware asynchronous schedule that differentiates semantically important keyframes from predictable non-keyframes. By assigning heterogeneous update frequencies and approximated trajectories to different frame groups, our method substantially reduces redundant computation while maintaining high visual fidelity.

Beyond its immediate efficiency gains, RhymeFlow reveals several promising directions for future exploration. One avenue is to replace the manually designed progressive schedule with learned, continuous scheduling functions that adapt to scene dynamics and model behavior. Moreover, the framework is inherently orthogonal to existing acceleration techniques such as sparse attention, quantization, and token pruning, enabling compounded improvements when combined. By shifting the focus from uniform denoising to intelligent, content-aware scheduling, RhymeFlow offers a principled path toward scalable and accessible high-resolution video generation.

\textbf{Acknowledgements.}
This work was supported in part by the Beijing Natural Science Foundation of China under Grant L252011, by the National Natural Science Foundation of China under Grant 62576185, and by the Young Elite Scientist Sponsorship Program by CAST under Grant YESS20240544.

\bibliographystyle{plain}
\bibliography{main}

\newpage
\appendix
\section{More Experimental Results}

\subsection{Additional Visualization Results}

\begin{figure*}[h]
  \centering
  \includegraphics[width=\textwidth]{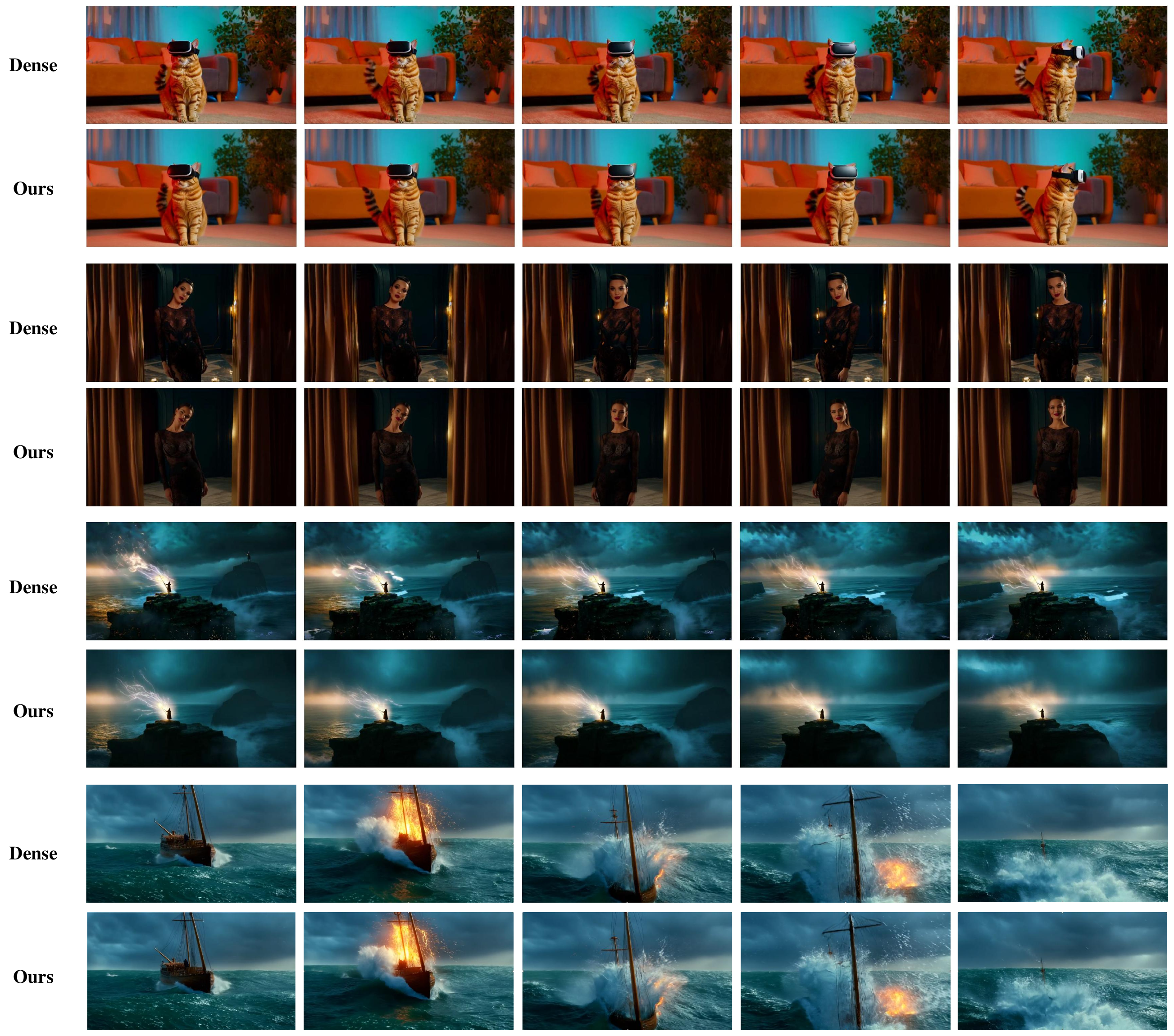}
  \caption{\textbf{Additional qualitative results on Wan 2.1.} We compare the sequences generated by the original model (utilizing full dense attention) against those produced by our RhymeFlow. Our method successfully preserves intricate details and complex motion dynamics without introducing noticeable artifacts.}
  \label{supfigtop}
\end{figure*}

We present further qualitative visual comparisons in Figure~\ref{supfigtop}. While image sequences offer a static perspective, the temporal coherence and motion dynamics are best observed in motion. Therefore, we encourage readers to review the extended video samples provided in the accompanying multimedia materials for a more comprehensive visual assessment of our approach.

\subsection{Evaluation on Long-Duration Videos}

While the main paper primarily evaluates models under the standard 81-frame context, we further extend our experiments to long-duration (240 frames) videos to rigorously assess the scalability and robustness of our framework. 

As demonstrated in Table~\ref{retable1}, RhymeFlow exhibits remarkable stability across various parameter configurations ($T_w$ and $M$), maintaining high structural similarity and perceptual quality. Furthermore, we benchmarked RhymeFlow against state-of-the-art (SOTA) training-free acceleration baselines on these extended video sequences. Detailed in Table~\ref{retable3}, our method and its synergistic variant (Ours+SVG) consistently achieve a superior trade-off, delivering up to $1.91 \times$ acceleration while preserving high visual fidelity compared to existing masking-based counterparts.

\begin{table}[t]
\centering
\caption{\textbf{Parameter sensitivity analysis on long-duration videos.} We ablate the warm-up duration ($T_w$) and the keyframe budget ($M$) on Wan 2.1.}
\label{retable1}
\begin{tabular*}{\linewidth}{@{\extracolsep{\fill}} lccccc }
\toprule
Method & PSNR $\uparrow$ & SSIM $\uparrow$ & LPIPS $\downarrow$ & Latency (s) $\downarrow$ & Speedup $\uparrow$\\
\midrule
Original & - & - & - &  5599 & - \\
\midrule
Ours ($T_w=10,M=15$)  & 25.122 & 0.806 & 0.251 & 3034  &  1.85 $\times$ \\
Ours ($T_w=10,M=21$)  & 25.169 & 0.809 & 0.247 & 3485  & 1.61 $\times$\\
Ours ($T_w=10,M=30$)  & 25.339 & 0.818 & 0.241 & 3959  & 1.41 $\times$\\
\midrule
Ours ($T_w=15,M=15$)  & 27.011 & 0.813 & 0.232 &  3467  & 1.61 $\times$  \\
\textbf{Ours} ($T_w=15,M=21$) & \textbf{27.256} & \textbf{0.822} & \textbf{0.224} &  \textbf{3626} &  \textbf{1.54 $\times$} \\
Ours ($T_w=15,M=30$)  & 27.832 & 0.840 & 0.210 & 4164  &  1.34 $\times$\\
\midrule
Ours ($T_w=20,M=15$)  & 27.585 & 0.857 & 0.205 &  3605  & 1.55 $\times$ \\
Ours ($T_w=20,M=21$)  & 27.597 & 0.858 & 0.203 &  3908 & 1.43 $\times$ \\
Ours ($T_w=20,M=30$)  & 28.009 & 0.863 & 0.198 & 4369  &  1.28 $\times$\\
\bottomrule
\end{tabular*} 
\end{table}

\begin{table}[t]
\centering
\caption{\textbf{Quantitative comparison against SOTA baseline methods on long-duration videos.} RhymeFlow achieves the optimal balance of efficiency and visual quality.}
\label{retable3}
{
\begin{tabular*}{\linewidth}{@{\extracolsep{\fill}} lccccc }
\toprule
Method & PSNR $\uparrow$ & SSIM $\uparrow$ & LPIPS $\downarrow$ & Latency (s) $\downarrow$ & Speedup $\uparrow$\\
\midrule
Original & - & - & - &  5599 & - \\
\midrule
SVG~\cite{SVG2025arXiv}   & 21.139 & 0.705 & 0.344 &  3943  &  1.42 $\times$ \\
SAP~\cite{SAP2025arXiv}  & 26.050 & 0.836 & 0.217 &  2978  & 1.88 $\times$\\
Ours  &  27.256 & 0.822 & 0.224 &  3626  &  1.54 $\times$\\
\textbf{Ours + SVG}  &  \textbf{26.585} & \textbf{0.842} & \textbf{0.211} & \textbf{2931}  & \textbf{1.91 $\times$} \\
\bottomrule
\end{tabular*}
}
\end{table}

\subsection{Ablation Study on Progressive Scheduling Parameters}

As detailed in the main paper, our progressive asynchronous scheduling mechanism relies on the update stride parameters ($n_{\text{small}}$ and $n_{\text{large}}$). We perform an ablation study on these settings using Wan 2.1, with the results summarized in Table~\ref{retable5}. The findings indicate a clear empirical trade-off: larger strides  prioritize inference speed over generation fidelity. This validates our default configuration ($n_{\text{small}}=2, n_{\text{large}}=3$) as the optimal balance for ensuring high visual quality while delivering substantial computational acceleration.

\begin{table}[t]
\centering                  
\caption{\textbf{Ablation study on the stride parameters of the progressive scheduling} ($T_w=15, M=7$).}
\label{retable5}
\begin{tabular*}{\linewidth}{@{\extracolsep{\fill}} lccccc }
\toprule
Method & PSNR $\uparrow$ & SSIM $\uparrow$ & LPIPS $\downarrow$ & Latency (s) $\downarrow$ & Speedup $\uparrow$\\
\midrule
Original & - & - & - &  960 & - \\
\midrule
\textbf{Ours} ($n_{\text{small}}=2,n_{\text{large}}=3$)  &   \textbf{27.287} & \textbf{0.814}   & \textbf{0.210}  &  \textbf{627}  &  \textbf{1.53 $\times$} \\
Ours ($n_{\text{small}}=2,n_{\text{large}}=4$)  & 25.178 & 0.804  &  0.251  &  596 & 1.61 $\times$\\
Ours ($n_{\text{small}}=2,n_{\text{large}}=5$)  & 24.459 & 0.787  & 0.268 &  568 & 1.69 $\times$\\
\midrule
Ours ($n_{\text{small}}=3,n_{\text{large}}=4$)  & 23.546 & 0.739  & 0.294  &  604  & 1.59 $\times$  \\
Ours ($n_{\text{small}}=3,n_{\text{large}}=5$)  & 21.995 & 0.724  & 0.322  &  561  & 1.71 $\times$ \\
Ours ($n_{\text{small}}=3,n_{\text{large}}=6$)  & 21.125 & 0.685  & 0.339  &  530  & 1.81 $\times$\\
\bottomrule
\end{tabular*}
\end{table}

\section{More Theoretical Analysis}
\label{sec:rationale}

\subsection{Theoretical Analysis on FLOPs}

In the main paper, we evaluate our acceleration effects primarily through experimental time consumption results. In this section, we provide a theoretical analysis of the speed-up ratio from the perspective of Floating Point Operations (FLOPs). Let us revisit our pipeline through the lens of FLOPs analysis.

\textbf{Model Parameters:}
\begin{itemize}
    \item $L$: Number of transformer layers
    \item $H$: Number of attention heads per layer
    \item $d$: Dimension of each attention head
    \item $D = H \times d$: Total hidden dimension
\end{itemize}

\textbf{Video Parameters:}
\begin{itemize}
    \item $F$: Number of video frames (after VAE temporal compression)
    \item $N$: Number of tokens per frame (i.e., spatial tokens)
    \item $S = F \times N$: Total sequence length per sample
    \item $C$: Context length (text prompt tokens, $C=0$ for Wan)
\end{itemize}

\textbf{Denoising Parameters:}
\begin{itemize}
    \item $T$: Total number of denoising steps
    \item $T_{\text{warmup}}$: Number of warmup steps with dense attention
    \item $M$: Number of identified keyframes ($M \ll F$)
    \item $n$: Skip interval for non-keyframes ($n \geq 2$)
    \item $T_{\text{RhymeFlow}} = T - T_{\text{warmup}}$: Number of RhymeFlow phase steps
\end{itemize}

Specifically, we use the Wan2.1-T2V-1.3B model configuration throughout our analysis. We now concentrate on the basic asynchronous scheduling implementations (without progressive scheduling), detailed statistics are illustrated in Table \ref{tab:wan_config}.

\begin{table}[h]
\centering
\caption{\textbf{Wan2.1 Model Configuration.}}
\label{tab:wan_config}
\begin{tabular}{lc}
\toprule
\textbf{Parameter} & \textbf{Value} \\
\midrule
Transformer layers ($L$) & 30 \\
Attention heads ($H$) & 12 \\
Head dimension ($d$) & 128 \\
Hidden dimension ($D$) & 1,536 \\
\midrule
Video resolution & 720p (1280$\times$720) \\
Original frames & 81 \\
Compressed frames ($F$) & 21 \\
Tokens per frame ($N$) & 3,600 \\
Total tokens ($S$) & 75,600 \\
\midrule
Total steps ($T$) & 50 \\
Warmup steps ($T_{\text{warmup}}$) & 10 \\
Keyframes ($M$) & 5 \\
Skip interval ($n$) & 2 \\
\bottomrule
\end{tabular}
\end{table}

\subsubsection{Dense Baseline FLOPS Analysis.}

We first analyze the computational cost of the standard dense attention mechanism, which serves as our baseline. A single self-attention layer at one denoising timestep involves the following operations:

\textbf{QKV Projection.} Transform input hidden states into query, key, and value representations:
\begin{equation}
\mathbf{Q} = \mathbf{X} \mathbf{W}_Q, \quad \mathbf{K} = \mathbf{X} \mathbf{W}_K, \quad \mathbf{V} = \mathbf{X} \mathbf{W}_V,
\end{equation}
where $\mathbf{X} \in \mathbb{R}^{S \times D}$ and $\mathbf{W}_Q, \mathbf{W}_K, \mathbf{W}_V \in \mathbb{R}^{D \times D}$. The total FLOPs for QKV projection:
\begin{equation}
\text{FLOPs}_{\text{QKV}} = 3 \times (S \times D \times D) = 3SD^2.
\end{equation}

\textbf{Attention Score Computation.} Compute pairwise attention scores $\mathbf{A} = \mathbf{Q}\mathbf{K}^\top / \sqrt{d}$:
\begin{equation}
\text{FLOPs}_{\text{scores}} = S \times S \times D = S^2D.
\end{equation}

\textbf{Softmax Normalization.} Apply softmax to obtain attention weights. While softmax involves exponentials and normalization, for FLOP counting we approximate this as:
\begin{equation}
\text{FLOPs}_{\text{softmax}} \approx 3S^2,
\end{equation}
which is typically negligible compared to matrix multiplications for large $D$.

\textbf{Attention-Weighted Aggregation.} Compute output as $\mathbf{O} = \text{softmax}(\mathbf{A})\mathbf{V}$:
\begin{equation}
\text{FLOPs}_{\text{aggregate}} = S \times S \times D = S^2D.
\end{equation}

\textbf{Output Projection.} Project concatenated multi-head outputs back to hidden dimension:
\begin{equation}
\text{FLOPs}_{\text{out}} = S \times D \times D = SD^2.
\end{equation}

Summing all components:
\begin{equation}
\begin{split}
\text{FLOPs}_{\text{layer}} &= 3SD^2 + S^2D + 3S^2 + S^2D + SD^2 \\
&= 4SD^2 + 2S^2D + 3S^2.
\end{split}
\end{equation}
For large-scale models where $S$ and $D$ are both large, the quadratic attention term $2S^2D$ typically dominates. We can approximate:
\begin{equation}
\text{FLOPs}_{\text{layer}} \approx 4SD^2 + 2S^2D = 2D(2SD + S^2).
\label{eq:flops_layer_approx}
\end{equation}
For the complete dense denoising process with $L$ layers and $T$ timesteps:
\begin{equation}
\text{FLOPs}_{\text{dense}} = L \times T \times \text{FLOPs}_{\text{layer}} = L \times T \times 2D(2SD + S^2).
\label{eq:flops_dense}
\end{equation}
Using the configuration in Table~\ref{tab:wan_config}:
\begin{align}
\text{FLOPs}_{\text{layer}} &= 4 \times 75{,}600 \times 1{,}536^2 + 2 \times 75{,}600^2 \times 1{,}536 \nonumber \\
&= 713{,}666{,}227{,}200 + 17{,}559{,}660{,}544{,}000 \nonumber \\
&\approx 18.27 \times 10^{12} \text{ FLOPs} = 18.27 \text{ TFLOPs}.
\end{align}
The attention core (quadratic term) accounts for:
\begin{equation}
\frac{2S^2D}{\text{FLOPs}_{\text{layer}}} = \frac{17.56}{18.27} \approx 96.1\%,
\end{equation}
confirming that attention dominates the computation.
Total dense baseline FLOPs:
\begin{equation}
\begin{split}
\text{FLOPs}_{\text{dense}} &= 30 \times 50 \times 18.27 = 27{,}405 \text{ TFLOPs} \\
&= 27.41 \text{ PFLOPs}.
\end{split}
\label{eq:flops_dense_wan}
\end{equation}

\subsubsection{RhymeFlow FLOPS Analysis.}

We now analyze the computational cost of our Selective Step Skipping method, which consists of two phases: warmup and selective denoising. 

\textbf{Warm up Stage.} During the warmup phase ($t \leq T_{\text{warmup}}$), all $F$ frames perform dense attention at every step to establish initial distributions. The FLOPs are identical to the dense baseline:
\begin{equation}
\text{FLOPs}_{\text{warmup}} = L \times T_{\text{warmup}} \times 2D(2SD + S^2).
\label{eq:flops_warmup}
\end{equation}
After warmup, we partition frames into keyframes $\mathcal{F}_k$ (with $|\mathcal{F}_k| = M$) and non-keyframes $\mathcal{F}_n$ (with $|\mathcal{F}_n| = F - M$). The denoising schedule is:
\begin{itemize}
    \item Keyframes ($f \in \mathcal{F}_k$): Denoise at every step
    \item Non-keyframes ($f \in \mathcal{F}_n$): Denoise only when $(t - T_{\text{warmup}}) \bmod n = 0$
\end{itemize}

\textbf{Skip Steps.} At timesteps where $(t - T_{\text{warmup}}) \bmod n \neq 0$, only keyframes compute attention outputs (non-keyframes use interpolated representations). The effective sequence length is:
\begin{equation}
S_{\text{skip}} = M \times N.
\end{equation}
Number of skip steps in RhymeFlow phase:
\begin{equation}
N_{\text{skip}} = \left\lfloor T_{\text{RhymeFlow}} \times \frac{n-1}{n} \right\rfloor.
\end{equation}
FLOPs for skip steps:
\begin{equation}
\begin{split}
    \text{FLOPs}_{\text{skip}} &= L \times N_{\text{skip}} \times 2D(2S_{\text{skip}}D + S_{\text{skip}}^2) \\ &= L \times N_{\text{skip}} \times 2D(2MND + M^2N^2).
\label{eq:flops_skip}
\end{split}
\end{equation}

\textbf{Full Denoising Steps.} At timesteps where $(t - T_{\text{warmup}}) \bmod n = 0$, both keyframes and non-keyframes denoise. All $F$ frames participate:
\begin{equation}
S_{\text{full}} = F \times N = S.
\end{equation}
Number of full denoising steps:
\begin{equation}
N_{\text{denoise}} = T_{\text{RhymeFlow}} - N_{\text{skip}} \approx \frac{T_{\text{RhymeFlow}}}{n}.
\end{equation}
FLOPs for full denoising steps:
\begin{equation}
\text{FLOPs}_{\text{denoise}} = L \times N_{\text{denoise}} \times 2D(2FND + F^2N^2).
\label{eq:flops_denoise}
\end{equation}
Combining both phases:
\begin{equation}
\text{FLOPs}_{\text{RhymeFlow}} = \text{FLOPs}_{\text{warmup}} + \text{FLOPs}_{\text{skip}} + \text{FLOPs}_{\text{denoise}}.
\label{eq:flops_RhymeFlow_total}
\end{equation}
Substituting Equations~\eqref{eq:flops_warmup}, \eqref{eq:flops_skip}, and \eqref{eq:flops_denoise}:
\begin{equation}
\begin{aligned}
\text{FLOPs}_{\text{RhymeFlow}} = &\ L \times T_{\text{warmup}} \times 2D(2FND + F^2N^2) \\
&+ L \times N_{\text{skip}} \times 2D(2MND + M^2N^2) \\
&+ L \times N_{\text{denoise}} \times 2D(2FND + F^2N^2).
\end{aligned}
\label{eq:flops_RhymeFlow_expanded}
\end{equation}

\subsubsection{Numerical Example: Wan 2.1}

Using $T=50$, $T_{\text{warmup}}=10$, $M=5$, $n=2$, $F=21$:
\begin{align}
T_{\text{RhymeFlow}} &= 50 - 10 = 40, \\
N_{\text{skip}} &= \left\lfloor 40 \times \frac{1}{2} \right\rfloor = 20, \\
N_{\text{denoise}} &= 40 - 20 = 20.
\end{align}

\textbf{Warmup FLOPs.}
\begin{equation}
\begin{split}
    \text{FLOPs}_{\text{warmup}} &= 30 \times 10 \times 18.27 = 5{,}481 \text{ TFLOPs} \\ &= 5.48 \text{ PFLOPs}.
\end{split}
\end{equation}

\textbf{Skip Steps FLOPs.}
First compute $\text{FLOPs}_{\text{layer, skip}}$ for sequence length $S_{\text{skip}} = 5 \times 3{,}600 = 18{,}000$:
\begin{align}
\text{FLOPs}_{\text{layer, skip}} &= 4 \times 18{,}000 \times 1{,}536^2 + 2 \times 18{,}000^2 \times 1{,}536 \nonumber \\
&= 170{,}074{,}521{,}600 + 1{,}990{,}656{,}000{,}000 \nonumber \\
&\approx 2.16 \times 10^{12} \text{ FLOPs} = 2.16 \text{ TFLOPs}.
\end{align}
Total skip FLOPs:
\begin{equation}
\text{FLOPs}_{\text{skip}} = 30 \times 20 \times 2.16 = 1{,}296 \text{ TFLOPs} = 1.30 \text{ PFLOPs}.
\end{equation}

\textbf{Denoise Steps FLOPs.}
\begin{equation}
\begin{split}
    \text{FLOPs}_{\text{denoise}} &= 30 \times 20 \times 18.27 = 10{,}962 \text{ TFLOPs} \\ &= 10.96 \text{ PFLOPs}.
\end{split}
\end{equation}

\textbf{Total RhymeFlow FLOPs.}
\begin{equation}
\text{FLOPs}_{\text{RhymeFlow}} = 5.48 + 1.30 + 10.96 = 17.74 \text{ PFLOPs}.
\label{eq:flops_RhymeFlow_wan_total}
\end{equation}

\subsubsection{Theoretical Speedup Analysis}

The theoretical speedup from FLOP reduction:
\begin{equation}
\text{Speedup}_{\text{FLOPs}} = \frac{\text{FLOPs}_{\text{dense}}}{\text{FLOPs}_{\text{RhymeFlow}}}.
\end{equation}
For Wan~2.1:
\begin{equation}
\text{Speedup}_{\text{FLOPs}} = \frac{27.41}{17.74} = 1.545\times.
\label{eq:speedup_flops_wan}
\end{equation}
This corresponds to:
\begin{equation}
\text{FLOP Reduction} = 1 - \frac{17.74}{27.41} = 35.3\%.
\end{equation}
The reported speedup ratio of $1.53 \times$ is slightly lower than the theoretical speedup of $1.545\times$. This discrepancy results from the non-ideal scaling of hardware efficiency and algorithmic overheads not captured by pure FLOPs counting. We attribute this gap to two primary factors:
\begin{itemize}
    \item \textbf{Algorithmic Overheads:} 
    The theoretical analysis assumes zero cost for control logic. However, the practical implementation of RhymeFlow introduces necessary computations:
    \begin{enumerate}
        \item Keyframe Identification: The computation of frame-to-frame latent similarity (e.g., cosine similarity) and the selection algorithm (clustering or thresholding) consume GPU cycles.
        \item Latent Trajectories Projection: Generating intermediate states ($x_{t-1}$) for skipped frames via flow-based latent projection requires additional vector operations, which, while lightweight, are not negligible.
    \end{enumerate}

    \item \textbf{Hardware Efficiency \& Memory Access Constraints:} 
    The reduction in FLOPs does not translate linearly to latency reduction due to decreased GPU utilization during the skip steps.
    \begin{enumerate}
        \item Reduced Parallelism: During skip steps, the model processes only $M=5$ keyframes instead of the full $F=21$ frames. This reduces the attention sequence length from $S_{\text{full}} = 75,600$ to $S_{\text{skip}} = 18,000$.
        \item GPU Occupancy Drop: On high-performance GPUs, such a significant reduction in sequence length ($\sim 76\%$ decrease) lowers the kernel occupancy. The workload shifts from being compute-bound to memory-bound, meaning the GPU cores spend more time waiting for data transfer than performing calculations. Consequently, the effective TFLOPs/s achieved during skip steps is lower than during dense full-sequence processing.
    \end{enumerate}
\end{itemize}

Therefore, the measured speedup of $1.53 \times$ represents a robust trade-off between theoretical reduction and hardware utilization efficiency.

\subsection{Theoretical Analysis on KV-Caching}

In the main paper, we introduce our KV-Caching mechanism in a simple manner. Now let's dive into the design of per-layer rolling-cache and analyze how efficient it is compared to the original KV management.
A key challenge in selective step skipping is maintaining temporal coherence for non-keyframes that do not perform denoising at certain timesteps. Traditional KV-caching mechanisms, widely adopted in autoregressive language models, exploit causal dependencies to reuse previously computed key-value pairs. However, these approaches are fundamentally inapplicable to video diffusion models due to three critical distinctions: (1)~non-autoregressive generation, where all frames are updated simultaneously at each denoising step; (2) bidirectional temporal dependencies, requiring full attention across all frames without causal constraints; and (3) state evolution, where the latent representations of all tokens change at every timestep.

To address this challenge, we propose a \textbf{rolling KV-cache strategy} that enables latent projection of intermediate frame states during skipped denoising steps. Our approach maintains temporal consistency while achieving substantial memory efficiency compared to naive caching of all intermediate states.

\subsubsection{Problem Formulation:}

Let $\mathbf{z}_t^{(f)} \in \mathbb{R}^{N \times D}$ denote the latent representation of frame $f$ at denoising timestep $t$, where $N$ is the number of tokens per frame and $D$ is the feature dimension. In the standard dense denoising schedule, every frame undergoes attention computation at every timestep:
\begin{equation}
\mathbf{z}_{t-1}^{(f)} = \text{Attention}\left(\mathbf{Q}_t^{(f)}, \mathbf{K}_t, \mathbf{V}_t\right) + \mathbf{z}_t^{(f)},
\end{equation}
where $\mathbf{K}_t, \mathbf{V}_t \in \mathbb{R}^{(F \cdot N) \times D}$ aggregate keys and values from all $F$ frames.

In our selective step skipping regime, we partition frames into keyframes $\mathcal{F}_k$ and non-keyframes $\mathcal{F}_n$. At non-denoising steps, non-keyframes must still provide key-value representations for keyframe attention, but their query outputs need not be computed. The key challenge is how we can obtain $\mathbf{K}_t^{(f)}$ and $\mathbf{V}_t^{(f)}$ for $f \in \mathcal{F}_n$ at skipped timesteps without performing full attention computation?

\subsubsection{Rolling Cache Design.}

We introduce a \textbf{per-layer, per-frame rolling cache} that stores the attention outputs of the two most recent denoising timesteps for each non-keyframe. For frame $f \in \mathcal{F}_n$ at layer $\ell$, the cache $\mathcal{C}_\ell^{(f)}$ maintains:
\begin{equation}
\mathcal{C}_\ell^{(f)} = \left\{
\begin{aligned}
&\mathbf{h}_{\text{before}}^{(\ell, f)} \in \mathbb{R}^{N \times D}, \quad &&t_{\text{before}}, \\
&\mathbf{h}_{\text{after}}^{(\ell, f)} \in \mathbb{R}^{N \times D}, \quad &&t_{\text{after}},
\end{aligned}
\right\}
\end{equation}
where $\mathbf{h}_{\text{before}}^{(\ell, f)}$ and $\mathbf{h}_{\text{after}}^{(\ell, f)}$ are the cached attention outputs at the two most recent denoising steps $t_{\text{before}} > t_{\text{after}}$ (noting that diffusion timesteps decrease during denoising). Crucially, we cache the \textbf{output hidden states} (post-attention) rather than input latents, as they already incorporate contextual information from all other frames.

At the end of the warmup phase (step $t_{\text{w}}$), we initialize the cache for all non-keyframes in all layers:
\begin{equation}
\begin{split}
    \mathbf{h}_{\text{before}}^{(\ell, f)} = \mathbf{h}_{\text{after}}^{(\ell, f)} = \text{Attention}^{(\ell)}\left(\mathbf{Q}_{t_{\text{warmup}}}^{(f)}, \mathbf{K}_{t_{\text{warmup}}}, \mathbf{V}_{t_{\text{warmup}}}\right), \\t_{\text{before}} = t_{\text{after}} = t_{\text{w}}.
\end{split}
\end{equation}


At a skipped timestep $t_{\text{skip}}$ where $t_{\text{after}} < t_{\text{skip}} < t_{\text{before}}$, we reconstruct the frame representation via latent trajectories projection:
\begin{equation}
\mathbf{h}_{t_{\text{skip}}}^{(\ell, f)} = (1 - \alpha) \cdot \mathbf{h}_{\text{before}}^{(\ell, f)} + \alpha \cdot \mathbf{h}_{\text{after}}^{(\ell, f)},
\end{equation}
where the interpolation weight $\alpha \in [0, 1]$ is computed based on the relative temporal position:
\begin{equation}
\alpha = \frac{t_{\text{before}} - t_{\text{skip}}}{t_{\text{before}} - t_{\text{after}} + \epsilon}, \quad \epsilon = 10^{-8}.
\end{equation}
This formulation ensures that $\alpha \to 0$ as $t_{\text{skip}} \to t_{\text{before}}$ (favoring the earlier cached state) and $\alpha \to 1$ as $t_{\text{skip}} \to t_{\text{after}}$ (favoring the later cached state), aligning with the decreasing-timestep denoising trajectory.

\textbf{Attention computation at skipped steps:} While non-keyframes $f \in \mathcal{F}_n$ do not compute query outputs, they must still provide key-value pairs for keyframe attention. We directly use the latent projected hidden states:
\begin{equation}
\mathbf{K}_{t_{\text{skip}}}^{(f)} = \mathbf{V}_{t_{\text{skip}}}^{(f)} = \mathbf{h}_{t_{\text{skip}}}^{(\ell, f)}, \quad f \in \mathcal{F}_n.
\end{equation}
Keyframes $f \in \mathcal{F}_k$ perform full attention over all frames:
\begin{equation}
\begin{split}
    \mathbf{h}_{t_{\text{skip}}}^{(\ell, f)} = \text{Attention}^{(\ell)} ( \mathbf{Q}_{t_{\text{skip}}}^{(f)}, \left[\mathbf{K}_{t_{\text{skip}}}^{(1)}, \ldots, \mathbf{K}_{t_{\text{skip}}}^{(F)}\right], \\ \left[\mathbf{V}_{t_{\text{skip}}}^{(1)}, \ldots, \mathbf{V}_{t_{\text{skip}}}^{(F)}\right]), \quad f \in \mathcal{F}_k.
\end{split}
\end{equation}

\textbf{Cache Update at Denoising Steps:} When non-keyframes perform denoising at "Rhythmic Point" $t_{\text{denoise}}$, we update the rolling cache via a shift-and-store strategy:

\begin{equation}
\begin{aligned}
\mathbf{h}_{\text{before}}^{(\ell, f)} &\leftarrow \mathbf{h}_{\text{after}}^{(\ell, f)}, \\
t_{\text{before}} &\leftarrow t_{\text{after}}, \\
\mathbf{h}_{\text{after}}^{(\ell, f)} &\leftarrow \text{Attn}^{(\ell)}\bigl(\mathbf{Q}_{t_{\text{denoise}}}^{(f)}, \mathbf{K}_{t_{\text{denoise}}}, \mathbf{V}_{t_{\text{denoise}}}\bigr), \\
t_{\text{after}} &\leftarrow t_{\text{denoise}}.
\end{aligned}
\end{equation}

This rolling update ensures that the cache always retains the two most recent denoising states, enabling accurate projection for the next $(n-1)$ skipped steps.

\subsubsection{Cross-Layer Consistency Guarantees.}

A subtle but critical challenge in multi-layer architectures is ensuring \textbf{temporal consistency} across layers. Inconsistent cache states can cause error propagation, as layer $\ell+1$ depends on the output of layer $\ell$. We enforce consistency through three mechanisms:

\begin{itemize}
\item \textbf{Shared global step counter}: A class-level variable is incremented only in layer~0 and shared across all layer instances, ensuring all layers agree on the current timestep.
\item \textbf{Synchronized cache updates}: All layers update their caches simultaneously at denoising steps, preventing temporal misalignment.
\item \textbf{Shared keyframe indices}: The set $\mathcal{F}_k$ is determined once at the end of warmup (layer~0) and frozen thereafter, ensuring all layers use the same denoising schedule.
\end{itemize}

Formally, let $\tau(\ell)$ denote the effective timestep used by layer $\ell$. Consistency requires $\tau(\ell) = \tau(\ell') = t_{\text{current}}$ for all $\ell, \ell'$, which our design guarantees.

\section{Detailed Implementation of RhymeFlow}
\label{sec:implementation}

To provide a clear procedural understanding of the proposed framework, we present the algorithmic details of RhymeFlow. The process is divided into two primary stages: (1) Content-aware sequential keyframe selection, and (2) Asynchronous denoising flow scheduling.

\subsection{Sequential Keyframe Selection}

As discussed in Section 3.1 of the main paper, traditional uniform frame selection fails to capture non-linear semantic transitions, while computing similarity on noisy latents $\mathbf{z}_T$ is unreliable due to noise interference. We first perform a single-step denoising proxy to estimate the clean latents $\hat{\mathbf{z}}_0$ as a robust basis for selection. To prevent keyframes from clustering too densely or being too sparse due to a fixed similarity threshold, we propose a Dynamic Threshold Adjustment mechanism, as detailed in Algorithm \ref{alg:keyframe_selection}. We define an expected average interval $L_{\text{avg}} = N/M$. If the temporal gap between the current and previous keyframe significantly exceeds $L_{\text{avg}}$ (controlled by $\delta_{up}$), the threshold $\tau$ is decreased by $\Delta \tau$ to encourage selection; conversely, if the gap is too small (controlled by $\delta_{\text{down}}$), $\tau$ is increased to suppress redundant keyframes. This ensures a content-adaptive yet well-distributed keyframe set.

\begin{algorithm}[t]
    \caption{Dynamic Sequential Keyframe Selection}
    \label{alg:keyframe_selection}
    \begin{algorithmic}[1]
        \Require Initial noisy latents $\{\mathbf{z}_{T}^{(i)}\}_{i=1}^{N}$, Initial threshold $\tau$, Max keyframe budget $M$, Threshold step $\Delta \tau$, Interval tolerance $\{\delta_{up}, \delta_{\text{down}}\}$.
        \Ensure Keyframe set $\mathcal{K}$, Non-keyframe set $\mathcal{N}$.
        \State \textit{// Step 1: Predict clean latent proxies}
        \State $\{\hat{\mathbf{z}}_0^{(i)}\}_{i=1}^{N} \gets \text{Denoise}(\{\mathbf{z}_{T}^{(i)}\}, T \to 0)$ \Comment{Single-step approximation}
        
        \State \textit{// Step 2: Adaptive sequential selection}
        \State $\mathcal{K} \gets \{1\}$, $idx_{\text{last}} \gets 1$, $L_{\text{avg}} \gets N/M$ 
        \For{$i = 2$ \textbf{to} $N$}
            \If{$|\mathcal{K}| < M$}
                \State $S \gets \text{CosineSimilarity}(\hat{\mathbf{z}}_0^{(i)}, \hat{\mathbf{z}}_0^{(idx_{\text{last}})})$
                \If{$S < \tau$}
                    \State $\Delta L \gets i - idx_{\text{last}}$ \Comment{Calculate actual interval}
                    \State \textit{// Dynamic threshold adjustment}
                    \If{$\Delta L \ge L_{\text{avg}} + \delta_{up}$} 
                        \State $\tau \gets \tau - \Delta \tau$ \Comment{Decrease threshold to relax selection}
                    \ElsIf{$\Delta L \le L_{\text{avg}} - \delta_{\text{down}}$}
                        \State $\tau \gets \tau + \Delta \tau$ \Comment{Increase threshold to tighten selection}
                    \EndIf
                    \State $\mathcal{K} \gets \mathcal{K} \cup \{i\}$
                    \State $idx_{\text{last}} \gets i$
                \EndIf
            \EndIf
        \EndFor
        \State $\mathcal{N} \gets \{1, \dots, N\} \setminus \mathcal{K}$
        \State \Return $\mathcal{K}$, $\mathcal{N}$
    \end{algorithmic}
\end{algorithm}

\subsection{The Overall RhymeFlow Pipeline}

The core of RhymeFlow lies in its heterogeneous scheduling. While keyframes maintain a standard step-by-step denoising trajectory to preserve structural integrity, non-keyframes follow an accelerated path by skipping redundant steps. Crucially, during skipped steps, we employ Latent Trajectory Projection to synthesize the missing temporal context, ensuring that keyframe updates remain globally coherent. This process is summarized in Algorithm \ref{alg:rhymeflow}.

\begin{algorithm}[t]
    \caption{RhymeFlow: Asynchronous Denoising Flow Scheduling}
    \label{alg:rhymeflow}
    \begin{algorithmic}[1]
        \Require Noisy latents $\{\mathbf{z}_{T}^{(i)}\}$, Timesteps $T$, Warm-up steps $T_w$, Mid-point $T_{\text{mid}}$, Progressive strides $\{n_{\text{small}}, n_{\text{large}}\}$.
        \Ensure Fully denoised latents $\{\mathbf{z}_0^{(i)}\}$.
        
        \State $\mathcal{K}, \mathcal{N} \gets \text{DynamicKeyframeSelection}(\dots)$ \Comment{Via Algorithm \ref{alg:keyframe_selection}}
        
        \For{$t = T$ \textbf{down to} $1$}
            \If{$t > T - T_w$} \Comment{\textbf{Phase I: Synchronous Warm-up}}
                \State $\{\mathbf{z}_{t-1}^{(i)}\}_{i=1}^{N_f} \gets \text{FullDenoiseStep}(\{\mathbf{z}_{t}^{(i)}\}_{i=1}^{N_f})$ \Comment{Full 3D Attention}
            \Else \Comment{\textbf{Phase II: Asynchronous Scheduling}}
                \State \textit{// Progressive Stride Assignment (Eq. 2)}
                \State $n_{\text{skip}} \gets n_{\text{small}} \ \text{if} \ t > T_{\text{mid}} \ \text{else} \ n_{\text{large}}$ 
                
                \State \textit{// Identify active frames to be updated}
                \State $\mathcal{I}_{\text{active}} \gets \mathcal{K} \cup \{j \in \mathcal{N} \mid (T - T_w - t) \equiv 0 \pmod{n_{\text{skip}}}\}$
                
                \For{each frame $i \in \mathcal{I}_{\text{active}}$}
                    \State \textit{// Latent Trajectory Projection (Sec 3.2)}
                    \State $\mathcal{C} \gets \{\mathbf{z}_t^{(k)}\}_{k \in \mathcal{K}} \cup \{\hat{\mathbf{z}}_t^{(j)}\}_{j \in \mathcal{N} \setminus \mathcal{I}_{\text{active}}}$ \Comment{Anchor + Projected states}
                    
                    \State $n_i \gets 1 \ \text{if} \ i \in \mathcal{K} \ \text{else} \ n_{\text{skip}}$ \Comment{Step size for key/non-keyframe}
                    \State $\mathbf{z}_{t-n_i}^{(i)} \gets \text{Denoise}(\mathbf{z}_t^{(i)} \mid \mathcal{C})$ \Comment{Heterogeneous update}
                \EndFor
                \State \textit{// Skipped non-keyframes $\{\mathbf{z}^{(j)}\}$ remain at state $t$ until next rhythmic point.}
            \EndIf
        \EndFor
        \State \Return $\{\mathbf{z}_0^{(i)}\}_{i=1}^{N_f}$
    \end{algorithmic}
\end{algorithm}

\section{Discussions}

\subsection{Orthogonality to Intra-step Sparsity}
\label{discussonorg}

A notable advantage of RhymeFlow is its orthogonality to existing intra-step sparse attention techniques. This orthogonality stems from operating on different dimensions of the computational workload. Methods like SVG~\cite{SVG2025arXiv} target \textbf{intra-step efficiency}, optimizing how tokens interact within a single attention computation, while our method addresses \textbf{inter-step efficiency}, determining which frames participate in the computation at each denoising step. This fundamental distinction allows the two approaches to be synergistically combined for enhanced acceleration performance.

To illustrate, consider an intra-step method like SAP that constructs a token-level sparse attention mask, $\mathbf{M}_{\text{SAP}}$. Concurrently, our framework generates a distinct frame-level mask, $\mathbf{M}_{\text{Rhyme}}$, which dictates the set of active and inactive frames for a given step. A hybrid, highly efficient attention mechanism can be formulated by computing the element-wise product of these two masks:

\begin{equation}
\mathbf{M}_{\text{combined}} = \mathbf{M}_{\text{Rhyme}} \odot \mathbf{M}_{\text{SAP}}
\label{eq:combined_mask}
\end{equation}

The resulting mask, $\mathbf{M}_{\text{combined}}$, imposes a hierarchical sparsity. It first prunes the extensive computations corresponding to skipped non-keyframes (as dictated by $\mathbf{M}_{\text{Rhyme}}$) and then further sparsifies the attention calculations among the remaining active frames based on the token-level patterns in $\mathbf{M}_{\text{SAP}}$. This fusion of inter-step and intra-step strategies unlocks compounded acceleration gains, paving the way for even more efficient video generation models.

\subsection{Analysis of Failure Cases}

\begin{figure}[h]
  \centering
  \includegraphics[width=1.0\linewidth]{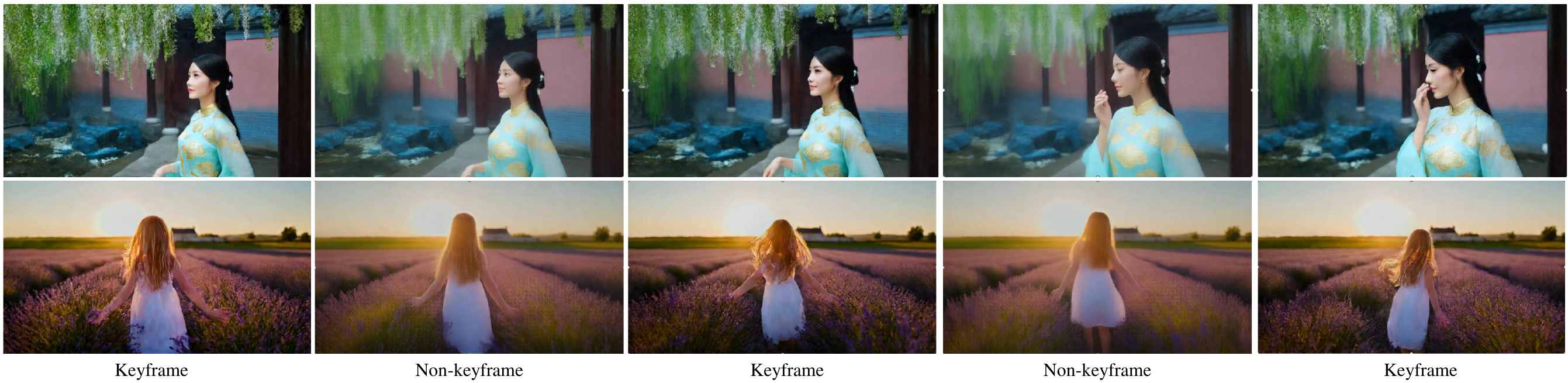}
   \caption{\textbf{Visual failure cases under extreme acceleration} ($n_{\text{skip}}=7$). Aggressive skipping leads to shimmering artifacts and loss of high-frequency textures in non-keyframes, highlighting the necessity of our progressive scheduling balance.}
   \label{fig:re2}
\end{figure}

To explore the upper bound of our acceleration framework, we conducted stress tests with an ultra-aggressive skipping stride ($n_{\text{skip}}=7$). As illustrated in Figure~\ref{fig:re2}, such extreme skipping triggers noticeable temporal aliasing and "shimmering" effects, particularly in non-keyframe regions. This degradation occurs because the linear approximation provided by our Latent Trajectory Projection assumes a locally smooth ODE path; at very large strides, this assumption fails to capture high-frequency motion non-linearities, causing asynchronous frames to drift from the anchor keyframes. These failure cases validate that our default conservative scheduling ($n_{\text{small}}=2, n_{\text{large}}=3$) is essential for maintaining artifact-free temporal coherence.

\subsection{Future Work}

Several promising avenues exist for extending our work. A primary direction is the exploration of learned asynchronous scheduling. While our current progressive strategy is heuristic-based, future research could employ a small policy network or uncertainty-based metric to dynamically determine which frames should be skipped and at what intervals, enabling a truly data-dependent acceleration.

Additionally, we envision the integration of RhymeFlow with post-training quantization (PTQ) and low-rank adaptations (LoRA). Since our method is strictly training-free and operates at the scheduling level, it remains orthogonal to model-level compression techniques. Combining these two paradigms could potentially unlock $3\times$ to $5\times$ speedups without compromising the generative capabilities of large-scale Video DiTs. Finally, investigating more sophisticated interpolation techniques beyond linear projection—such as utilizing optical flow priors—could further suppress artifacts at even larger skip strides.

\end{document}


\title{Supplementary Material for RhymeFlow: Training-Free Acceleration for Video Generation with Asynchronous Denoising Flow Scheduling} 

\titlerunning{Abbreviated paper title}

\author{First Author\inst{1}\orcidlink{0000-1111-2222-3333} \and
Second Author\inst{2,3}\orcidlink{1111-2222-3333-4444} \and
Third Author\inst{3}\orcidlink{2222--3333-4444-5555}}

\authorrunning{F.~Author et al.}

\institute{Princeton University, Princeton NJ 08544, USA \and
Springer Heidelberg, Tiergartenstr.~17, 69121 Heidelberg, Germany
\email{lncs@springer.com}\\
\url{http://www.springer.com/gp/computer-science/lncs} \and
ABC Institute, Rupert-Karls-University Heidelberg, Heidelberg, Germany\\
\email{\{abc,lncs\}@uni-heidelberg.de}}

\maketitle

\appendix

\section{More Experimental Results}

\subsection{Additional Visualization Results}

\begin{figure*}[t]
  \centering
  \includegraphics[width=\textwidth]{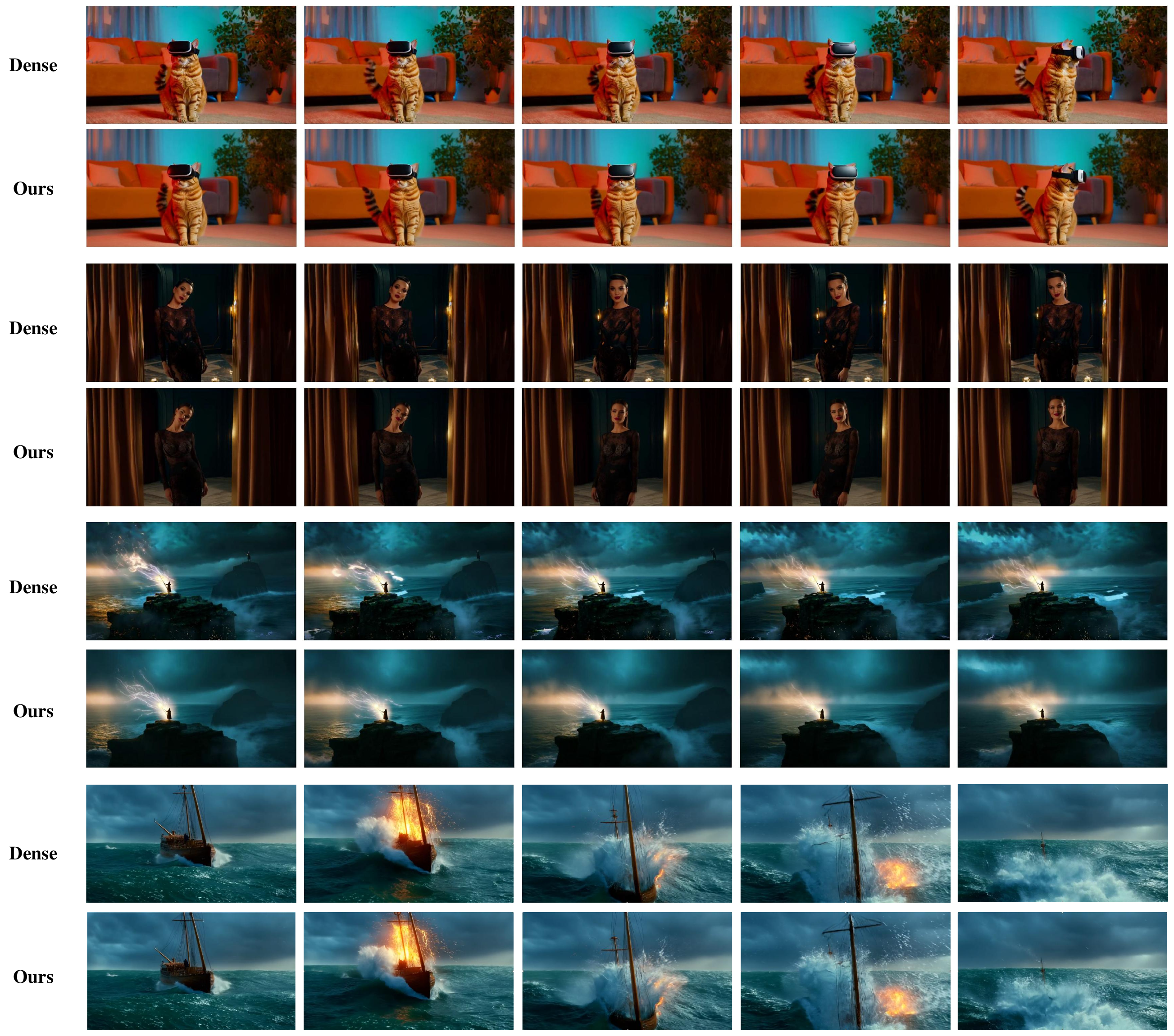}
  \caption{Additional qualitative results on Wan 2.1. We compare the sequences generated by the original model (utilizing full dense attention) against those produced by our \textit{RhymeFlow}. Our method successfully preserves intricate details and complex motion dynamics without introducing noticeable artifacts.}
  \label{supfigtop}
\end{figure*}

We present further qualitative visual comparisons in Figure~\ref{supfigtop}. While image sequences offer a static perspective, the temporal coherence and motion dynamics are best observed in motion. Therefore, we encourage readers to review the extended video samples provided in the accompanying multimedia materials for a more comprehensive visual assessment of our approach.

\subsection{Evaluation on Long-Duration  Videos}

While the main paper primarily evaluates models under the standard 81-frame context, we further extend our experiments to long-duration (240 frames) videos to rigorously assess the scalability and robustness of our framework. 

As demonstrated in Table~\ref{retable1}, \textit{RhymeFlow} exhibits remarkable stability across various parameter configurations ($T_w$ and $M$), maintaining high structural similarity and perceptual quality. Furthermore, we benchmarked \textit{RhymeFlow} against state-of-the-art (SOTA) training-free acceleration baselines on these extended video sequences. Detailed in Table~\ref{retable3}, our method and its synergistic variant (\textit{Ours+SVG}) consistently achieve a superior trade-off, delivering up to $1.91 \times$ acceleration while preserving high visual fidelity compared to existing masking-based counterparts.

\begin{table}[t]
\centering
\caption{Parameter sensitivity analysis on long-duration (240 frames) videos. We ablate the warm-up duration ($T_w$) and the keyframe budget ($M$) on Wan 2.1.}
\label{retable1}
\begin{tabular*}{\linewidth}{@{\extracolsep{\fill}} lccccc }
\toprule
Method & PSNR $\uparrow$ & SSIM $\uparrow$ & LPIPS $\downarrow$ & Latency (s) $\downarrow$ & Speedup $\uparrow$\\
\midrule
Original & - & - & - &  5599 & - \\
\midrule
Ours ($T_w=10,M=15$)  & 25.122 & 0.806 & 0.251 & 3034  &  1.85 $\times$ \\
Ours ($T_w=10,M=21$)  & 25.169 & 0.809 & 0.247 & 3485  & 1.61 $\times$\\
Ours ($T_w=10,M=30$)  & 25.339 & 0.818 & 0.241 & 3959  & 1.41 $\times$\\
\midrule
Ours ($T_w=15,M=15$)  & 27.011 & 0.813 & 0.232 &  3467  & 1.61 $\times$  \\
\textbf{Ours ($T_w=15,M=21$)}  & \textbf{27.256} & \textbf{0.822} & \textbf{0.224} &  \textbf{3626} &  \textbf{1.54 $\times$} \\
Ours ($T_w=15,M=30$)  & 27.832 & 0.840 & 0.210 & 4164  &  1.34 $\times$\\
\midrule
Ours ($T_w=20,M=15$)  & 27.585 & 0.857 & 0.205 &  3605  & 1.55 $\times$ \\
Ours ($T_w=20,M=21$)  & 27.597 & 0.858 & 0.203 &  3908 & 1.43 $\times$ \\
Ours ($T_w=20,M=30$)  & 28.009 & 0.863 & 0.198 & 4369  &  1.28 $\times$\\
\bottomrule
\end{tabular*} 
\end{table}

\begin{table}[t]
\centering
\caption{Quantitative comparison against SOTA baseline methods on long-duration videos. \textit{RhymeFlow} achieves the optimal balance of efficiency and visual quality.}
\label{retable3}
{
\begin{tabular*}{\linewidth}{@{\extracolsep{\fill}} lccccc }
\toprule
Method & PSNR $\uparrow$ & SSIM $\uparrow$ & LPIPS $\downarrow$ & Latency (s) $\downarrow$ & Speedup $\uparrow$\\
\midrule
Original & - & - & - &  5599 & - \\
\midrule
SVG ~\cite{SVG2025arXiv}   & 21.139 & 0.705 & 0.344 &  3943  &  1.42 $\times$ \\
SAP ~\cite{SAP2025arXiv}  & 26.050 & 0.836 & 0.217 &  2978  & 1.88 $\times$\\
Ours  &  27.256 & 0.822 & 0.224 &  3626  &  1.54 $\times$\\
\textbf{Ours + SVG}  &  \textbf{26.585} & \textbf{0.842} & \textbf{0.211} & \textbf{2931}  & \textbf{1.91 $\times$} \\
\bottomrule
\end{tabular*}
}
\end{table}

\subsection{Ablation Study on Progressive Scheduling Parameters}

As detailed in the main paper, our progressive asynchronous scheduling mechanism relies on the update stride parameters ($n_{small}$ and $n_{large}$). We perform an ablation study on these settings using Wan 2.1, with the results summarized in Table~\ref{retable5}. The findings indicate a clear empirical trade-off: larger strides  prioritize inference speed over generation fidelity. This validates our default configuration ($n_{small}=2, n_{large}=3$) as the optimal balance for ensuring high visual quality while delivering substantial computational acceleration.

\begin{table}[t]
\centering                  
\caption{Ablation study on the stride parameters of the progressive scheduling ($T_w=15, M=7$).}
\label{retable5}
\begin{tabular*}{\linewidth}{@{\extracolsep{\fill}} lccccc }
\toprule
Method & PSNR $\uparrow$ & SSIM $\uparrow$ & LPIPS $\downarrow$ & Latency (s) $\downarrow$ & Speedup $\uparrow$\\
\midrule
Original & - & - & - &  960 & - \\
\midrule
\textbf{Ours ($n_{small}=2,n_{large}=3$)}  &   \textbf{27.287} & \textbf{0.814}   & \textbf{0.210}  &  \textbf{627}  &  \textbf{1.53 $\times$} \\
Ours ($n_{small}=2,n_{large}=4$)  & 25.178 & 0.804  &  0.251  &  596 & 1.61 $\times$\\
Ours ($n_{small}=2,n_{large}=5$)  & 24.459 & 0.787  & 0.268 &  568 & 1.69 $\times$\\
\midrule
Ours ($n_{small}=3,n_{large}=4$)  & 23.546 & 0.739  & 0.294  &  604  & 1.59 $\times$  \\
Ours ($n_{small}=3,n_{large}=5$)  & 21.995 & 0.724  & 0.322  &  561  & 1.71 $\times$ \\
Ours ($n_{small}=3,n_{large}=6$)  & 21.125 & 0.685  & 0.339  &  530  & 1.81 $\times$\\
\bottomrule
\end{tabular*}
\end{table}

\subsection{Subjective Evaluation (User Study)}

To rigorously validate the perceptual fidelity of \textit{RhymeFlow} from a human-centric perspective, we conducted a double-blind A/B test. Twenty participants evaluated the output videos conditioned on 20 diverse and challenging prompts. We benchmarked \textit{RhymeFlow} against two SOTA acceleration baselines (SVG ~\cite{SVG} and SAP ~\cite{SAP2025arXiv}), as well as the original model (Wan 2.1). 

The results, presented in Table~\ref{retab:user_study}, demonstrate that \textit{RhymeFlow} significantly outperforms intra-step sparse-attention baselines. Notably, in terms of Temporal Coherence, our method secured a 54.2\% win rate against SVG, confirming that our Latent Trajectory Projection effectively mitigates the motion jitter and incoherence commonly caused by standard masking-based methods. Crucially, when compared against the Original model, evaluators overwhelmingly reported a ``Tie'' across all metrics (ranging from 62.5\% to 66.6\%), underscoring that \textit{RhymeFlow} delivers a $1.53\times$ speedup with negligible perceptual degradation to human eyes.

\begin{table}[t]
\centering
\caption{Results of the double-blind user study. Values indicate the percentage of user preferences across pairwise comparisons.}
\label{retab:user_study}
{%
\begin{tabular}{llccc}
\toprule
\textbf{Comparison} & \textbf{Evaluation Metric} & \textbf{Ours Preferred} & \textbf{Tie} & \textbf{Competitor Preferred} \\
\midrule
\multirow{3}{*}{\shortstack[l]{Ours vs. \\ SVG}} 
 & Visual Quality & \textbf{45.8\%} & 37.5\% & 16.7\% \\
 & Temporal Coherence & \textbf{54.2\%} & 33.3\% & 12.5\% \\
 & Perceptual Preference & \textbf{50.0\%} & 37.5\% & 12.5\% \\
\midrule
\multirow{3}{*}{\shortstack[l]{Ours vs. \\ SAP}} 
 & Visual Quality & 37.5\% & \textbf{45.8\%} & 16.7\% \\
 & Temporal Coherence & \textbf{45.9\%} & 33.3\% & 20.8\% \\
 & Perceptual Preference & \textbf{45.8\%} & 41.7\% & 12.5\% \\
\midrule
\multirow{3}{*}{\shortstack[l]{ Ours vs. \\Original}} 
 & Visual Quality & 12.5\% & \textbf{62.5\%} & 25.0\% \\
 & Temporal Coherence & 16.7\% & \textbf{66.6\%} & 16.7\% \\
 & Perceptual Preference & 12.5\% & \textbf{62.5\%} & 25.0\% \\
\bottomrule
\end{tabular}
}
\end{table}


\section{More Theoretical Analysis}
\label{sec:rationale}

\subsection{Theoretical Analysis on FLOPs}

In the main paper, we evaluate our acceleration effects primarily through experimental time consumption results. In this section, we provide a theoretical analysis of the speed-up ratio from the perspective of Floating Point Operations (FLOPs). Let us revisit our pipeline through the lens of FLOPs analysis.

\paragraph{Model Parameters:}
\begin{itemize}
    \item $L$: Number of transformer layers
    \item $H$: Number of attention heads per layer
    \item $d$: Dimension of each attention head
    \item $D = H \times d$: Total hidden dimension
\end{itemize}
\paragraph{Video Parameters:}
\begin{itemize}
    \item $F$: Number of video frames (after VAE temporal compression)
    \item $N$: Number of tokens per frame (i.e., spatial tokens)
    \item $S = F \times N$: Total sequence length per sample
    \item $C$: Context length (text prompt tokens, $C=0$ for Wan)
\end{itemize}
\paragraph{Denoising Parameters:}
\begin{itemize}
    \item $T$: Total number of denoising steps
    \item $T_{\text{warmup}}$: Number of warmup steps with dense attention
    \item $M$: Number of identified keyframes ($M \ll F$)
    \item $n$: Skip interval for non-keyframes ($n \geq 2$)
    \item $T_{\text{RhymeFlow}} = T - T_{\text{warmup}}$: Number of RhymeFlow phase steps
\end{itemize}

Specifically, we use the Wan2.1-T2V-1.3B model configuration throughout our analysis. We now concentrate on the basic asynchronous scheduling implementations (without progressive scheduling), detailed statistics are illustrated in Table \ref{tab:wan_config}.

\begin{table}[h]
\centering
\caption{Wan2.1 Model Configuration}
\label{tab:wan_config}
\begin{tabular}{lc}
\toprule
\textbf{Parameter} & \textbf{Value} \\
\midrule
Transformer layers ($L$) & 30 \\
Attention heads ($H$) & 12 \\
Head dimension ($d$) & 128 \\
Hidden dimension ($D$) & 1,536 \\
\midrule
Video resolution & 720p (1280$\times$720) \\
Original frames & 81 \\
Compressed frames ($F$) & 21 \\
Tokens per frame ($N$) & 3,600 \\
Total tokens ($S$) & 75,600 \\
\midrule
Total steps ($T$) & 50 \\
Warmup steps ($T_{\text{warmup}}$) & 10 \\
Keyframes ($M$) & 5 \\
Skip interval ($n$) & 2 \\
\bottomrule
\end{tabular}
\end{table}

\subsubsection{Dense Baseline FLOPS Analysis.}

We first analyze the computational cost of the standard dense attention mechanism, which serves as our baseline. A single self-attention layer at one denoising timestep involves the following operations:

\paragraph{QKV Projection.} Transform input hidden states into query, key, and value representations:
\begin{equation}
\mathbf{Q} = \mathbf{X} \mathbf{W}_Q, \quad \mathbf{K} = \mathbf{X} \mathbf{W}_K, \quad \mathbf{V} = \mathbf{X} \mathbf{W}_V,
\end{equation}
where $\mathbf{X} \in \mathbb{R}^{S \times D}$ and $\mathbf{W}_Q, \mathbf{W}_K, \mathbf{W}_V \in \mathbb{R}^{D \times D}$. The total FLOPs for QKV projection:
\begin{equation}
\text{FLOPs}_{\text{QKV}} = 3 \times (S \times D \times D) = 3SD^2.
\end{equation}

\paragraph{Attention Score Computation.} Compute pairwise attention scores $\mathbf{A} = \mathbf{Q}\mathbf{K}^\top / \sqrt{d}$:
\begin{equation}
\text{FLOPs}_{\text{scores}} = S \times S \times D = S^2D.
\end{equation}

\paragraph{Softmax Normalization.} Apply softmax to obtain attention weights. While softmax involves exponentials and normalization, for FLOP counting we approximate this as:
\begin{equation}
\text{FLOPs}_{\text{softmax}} \approx 3S^2,
\end{equation}
which is typically negligible compared to matrix multiplications for large $D$.

\paragraph{Attention-Weighted Aggregation.} Compute output as $\mathbf{O} = \text{softmax}(\mathbf{A})\mathbf{V}$:
\begin{equation}
\text{FLOPs}_{\text{aggregate}} = S \times S \times D = S^2D.
\end{equation}

\paragraph{Output Projection.} Project concatenated multi-head outputs back to hidden dimension:
\begin{equation}
\text{FLOPs}_{\text{out}} = S \times D \times D = SD^2.
\end{equation}

\paragraph{Total FLOPs per Layer per Step.} Summing all components:
\begin{equation}
\begin{split}
\text{FLOPs}_{\text{layer}} &= 3SD^2 + S^2D + 3S^2 + S^2D + SD^2 \\
&= 4SD^2 + 2S^2D + 3S^2.
\end{split}
\end{equation}

For large-scale models where $S$ and $D$ are both large, the quadratic attention term $2S^2D$ typically dominates. We can approximate:
\begin{equation}
\text{FLOPs}_{\text{layer}} \approx 4SD^2 + 2S^2D = 2D(2SD + S^2).
\label{eq:flops_layer_approx}
\end{equation}

For the complete dense denoising process with $L$ layers and $T$ timesteps:
\begin{equation}
\text{FLOPs}_{\text{dense}} = L \times T \times \text{FLOPs}_{\text{layer}} = L \times T \times 2D(2SD + S^2).
\label{eq:flops_dense}
\end{equation}

Using the configuration in Table~\ref{tab:wan_config}:
\begin{align}
\text{FLOPs}_{\text{layer}} &= 4 \times 75{,}600 \times 1{,}536^2 + 2 \times 75{,}600^2 \times 1{,}536 \nonumber \\
&= 713{,}666{,}227{,}200 + 17{,}559{,}660{,}544{,}000 \nonumber \\
&\approx 18.27 \times 10^{12} \text{ FLOPs} = 18.27 \text{ TFLOPs}.
\end{align}

The attention core (quadratic term) accounts for:
\begin{equation}
\frac{2S^2D}{\text{FLOPs}_{\text{layer}}} = \frac{17.56}{18.27} \approx 96.1\%,
\end{equation}
confirming that attention dominates the computation.

Total dense baseline FLOPs:
\begin{equation}
\begin{split}
\text{FLOPs}_{\text{dense}} &= 30 \times 50 \times 18.27 = 27{,}405 \text{ TFLOPs} \\
&= 27.41 \text{ PFLOPs}.
\end{split}
\label{eq:flops_dense_wan}
\end{equation}

\subsubsection{RhymeFlow FLOPS Analysis.}

We now analyze the computational cost of our Selective Step Skipping method, which consists of two phases: warmup and selective denoising. 

\paragraph{Warm up Stage.} During the warmup phase ($t \leq T_{\text{warmup}}$), all $F$ frames perform dense attention at every step to establish initial distributions. The FLOPs are identical to the dense baseline:
\begin{equation}
\text{FLOPs}_{\text{warmup}} = L \times T_{\text{warmup}} \times 2D(2SD + S^2).
\label{eq:flops_warmup}
\end{equation}

After warmup, we partition frames into keyframes $\mathcal{F}_k$ (with $|\mathcal{F}_k| = M$) and non-keyframes $\mathcal{F}_n$ (with $|\mathcal{F}_n| = F - M$). The denoising schedule is:

\begin{itemize}
    \item \textbf{Keyframes} ($f \in \mathcal{F}_k$): Denoise at \textit{every} step
    \item \textbf{Non-keyframes} ($f \in \mathcal{F}_n$): Denoise only when $(t - T_{\text{warmup}}) \bmod n = 0$
\end{itemize}

This creates two types of steps:

\paragraph{Skip Steps.} At timesteps where $(t - T_{\text{warmup}}) \bmod n \neq 0$, only keyframes compute attention outputs (non-keyframes use interpolated representations). The effective sequence length is:
\begin{equation}
S_{\text{skip}} = M \times N.
\end{equation}

Number of skip steps in RhymeFlow phase:
\begin{equation}
N_{\text{skip}} = \left\lfloor T_{\text{RhymeFlow}} \times \frac{n-1}{n} \right\rfloor.
\end{equation}

FLOPs for skip steps:
\begin{equation}
\begin{split}
    \text{FLOPs}_{\text{skip}} &= L \times N_{\text{skip}} \times 2D(2S_{\text{skip}}D + S_{\text{skip}}^2) \\ &= L \times N_{\text{skip}} \times 2D(2MND + M^2N^2).
\label{eq:flops_skip}
\end{split}
\end{equation}

\paragraph{Full Denoising Steps.} At timesteps where $(t - T_{\text{warmup}}) \bmod n = 0$, both keyframes and non-keyframes denoise. All $F$ frames participate:
\begin{equation}
S_{\text{full}} = F \times N = S.
\end{equation}

Number of full denoising steps:
\begin{equation}
N_{\text{denoise}} = T_{\text{RhymeFlow}} - N_{\text{skip}} \approx \frac{T_{\text{RhymeFlow}}}{n}.
\end{equation}

FLOPs for full denoising steps:
\begin{equation}
\text{FLOPs}_{\text{denoise}} = L \times N_{\text{denoise}} \times 2D(2FND + F^2N^2).
\label{eq:flops_denoise}
\end{equation}

Combining both phases:
\begin{equation}
\text{FLOPs}_{\text{RhymeFlow}} = \text{FLOPs}_{\text{warmup}} + \text{FLOPs}_{\text{skip}} + \text{FLOPs}_{\text{denoise}}.
\label{eq:flops_RhymeFlow_total}
\end{equation}

Substituting Equations~\eqref{eq:flops_warmup}, \eqref{eq:flops_skip}, and \eqref{eq:flops_denoise}:
\begin{equation}
\begin{aligned}
\text{FLOPs}_{\text{RhymeFlow}} = &\ L \times T_{\text{warmup}} \times 2D(2FND + F^2N^2) \\
&+ L \times N_{\text{skip}} \times 2D(2MND + M^2N^2) \\
&+ L \times N_{\text{denoise}} \times 2D(2FND + F^2N^2).
\end{aligned}
\label{eq:flops_RhymeFlow_expanded}
\end{equation}

\subsubsection{Numerical Example: Wan 2.1:}

Using $T=50$, $T_{\text{warmup}}=10$, $M=5$, $n=2$, $F=21$:

\begin{align}
T_{\text{RhymeFlow}} &= 50 - 10 = 40, \\
N_{\text{skip}} &= \left\lfloor 40 \times \frac{1}{2} \right\rfloor = 20, \\
N_{\text{denoise}} &= 40 - 20 = 20.
\end{align}

\paragraph{Warmup FLOPs:}
\begin{equation}
\begin{split}
    \text{FLOPs}_{\text{warmup}} &= 30 \times 10 \times 18.27 = 5{,}481 \text{ TFLOPs} \\ &= 5.48 \text{ PFLOPs}.
\end{split}
\end{equation}

\paragraph{Skip Steps FLOPs:}
First compute $\text{FLOPs}_{\text{layer, skip}}$ for sequence length $S_{\text{skip}} = 5 \times 3{,}600 = 18{,}000$:
\begin{align}
\text{FLOPs}_{\text{layer, skip}} &= 4 \times 18{,}000 \times 1{,}536^2 + 2 \times 18{,}000^2 \times 1{,}536 \nonumber \\
&= 170{,}074{,}521{,}600 + 1{,}990{,}656{,}000{,}000 \nonumber \\
&\approx 2.16 \times 10^{12} \text{ FLOPs} = 2.16 \text{ TFLOPs}.
\end{align}

Total skip FLOPs:
\begin{equation}
\text{FLOPs}_{\text{skip}} = 30 \times 20 \times 2.16 = 1{,}296 \text{ TFLOPs} = 1.30 \text{ PFLOPs}.
\end{equation}

\paragraph{Denoise Steps FLOPs:}
\begin{equation}
\begin{split}
    \text{FLOPs}_{\text{denoise}} &= 30 \times 20 \times 18.27 = 10{,}962 \text{ TFLOPs} \\ &= 10.96 \text{ PFLOPs}.
\end{split}
\end{equation}

\paragraph{Total RhymeFlow FLOPs:}
\begin{equation}
\text{FLOPs}_{\text{RhymeFlow}} = 5.48 + 1.30 + 10.96 = 17.74 \text{ PFLOPs}.
\label{eq:flops_RhymeFlow_wan_total}
\end{equation}

\subsubsection{Theoretical Speedup Analysis}

The theoretical speedup from FLOP reduction:
\begin{equation}
\text{Speedup}_{\text{FLOPs}} = \frac{\text{FLOPs}_{\text{dense}}}{\text{FLOPs}_{\text{RhymeFlow}}}.
\end{equation}

For Wan~2.1:
\begin{equation}
\text{Speedup}_{\text{FLOPs}} = \frac{27.41}{17.74} = 1.545\times.
\label{eq:speedup_flops_wan}
\end{equation}

This corresponds to:
\begin{equation}
\text{FLOP Reduction} = 1 - \frac{17.74}{27.41} = 35.3\%.
\end{equation}

The reported speedup ratio of $1.53 \times$ is slightly lower than the theoretical speedup of $1.545\times$. This discrepancy results from the non-ideal scaling of hardware efficiency and algorithmic overheads not captured by pure FLOPs counting. We attribute this gap to two primary factors:

\begin{itemize}
    \item \textbf{Algorithmic Overheads:} 
    The theoretical analysis assumes zero cost for control logic. However, the practical implementation of RhymeFlow introduces necessary computations:
    \begin{enumerate}
        \item \textit{Keyframe Identification:} The computation of frame-to-frame latent similarity (e.g., cosine similarity) and the selection algorithm (clustering or thresholding) consume GPU cycles.
        \item \textit{Latent Trajectories Projection:} Generating intermediate states ($x_{t-1}$) for skipped frames via flow-based latent projection requires additional vector operations, which, while lightweight, are not negligible.
    \end{enumerate}

    \item \textbf{Hardware Efficiency \& Memory Access Constraints:} 
    The reduction in FLOPs does not translate linearly to latency reduction due to decreased GPU utilization during the \textit{skip steps}.
    \begin{enumerate}
        \item \textit{Reduced Parallelism:} During skip steps, the model processes only $M=5$ keyframes instead of the full $F=21$ frames. This reduces the attention sequence length from $S_{\text{full}} = 75,600$ to $S_{\text{skip}} = 18,000$.
        \item \textit{GPU Occupancy Drop:} On high-performance GPUs, such a significant reduction in sequence length ($\sim 76\%$ decrease) lowers the kernel occupancy. The workload shifts from being compute-bound to memory-bound, meaning the GPU cores spend more time waiting for data transfer than performing calculations. Consequently, the effective TFLOPs/s achieved during skip steps is lower than during dense full-sequence processing.
    \end{enumerate}
\end{itemize}

Therefore, the measured speedup of $1.53 \times$ represents a robust trade-off between theoretical reduction and hardware utilization efficiency.

\subsection{Theoretical Analysis on KV-Caching}

In the main paper, we introduce our KV-Caching mechanism in a simple manner. Now let's dive into the design of per-layer rolling-cache and analyze how efficient it is compared to the original KV management.
A key challenge in selective step skipping is maintaining temporal coherence for non-keyframes that do not perform denoising at certain timesteps. Traditional KV-caching mechanisms, widely adopted in autoregressive language models, exploit causal dependencies to reuse previously computed key-value pairs. However, these approaches are fundamentally inapplicable to video diffusion models due to three critical distinctions: (1)~\textit{non-autoregressive generation}, where all frames are updated simultaneously at each denoising step; (2)~\textit{bidirectional temporal dependencies}, requiring full attention across all frames without causal constraints; and (3)~\textit{state evolution}, where the latent representations of all tokens change at every timestep.

To address this challenge, we propose a \textit{rolling KV-cache strategy} that enables latent projection of intermediate frame states during skipped denoising steps. Our approach maintains temporal consistency while achieving substantial memory efficiency compared to naive caching of all intermediate states.

\subsubsection{Problem Formulation:}

Let $\mathbf{z}_t^{(f)} \in \mathbb{R}^{N \times D}$ denote the latent representation of frame $f$ at denoising timestep $t$, where $N$ is the number of tokens per frame and $D$ is the feature dimension. In the standard dense denoising schedule, every frame undergoes attention computation at every timestep:
\begin{equation}
\mathbf{z}_{t-1}^{(f)} = \text{Attention}\left(\mathbf{Q}_t^{(f)}, \mathbf{K}_t, \mathbf{V}_t\right) + \mathbf{z}_t^{(f)},
\end{equation}
where $\mathbf{K}_t, \mathbf{V}_t \in \mathbb{R}^{(F \cdot N) \times D}$ aggregate keys and values from all $F$ frames.

In our selective step skipping regime, we partition frames into keyframes $\mathcal{F}_k$ and non-keyframes $\mathcal{F}_n$. At non-denoising steps, non-keyframes must still provide key-value representations for keyframe attention, but their query outputs need not be computed. The key challenge is how we can obtain $\mathbf{K}_t^{(f)}$ and $\mathbf{V}_t^{(f)}$ for $f \in \mathcal{F}_n$ at skipped timesteps without performing full attention computation?

\subsubsection{Rolling Cache Design.}

We introduce a \textbf{per-layer, per-frame rolling cache} that stores the attention outputs of the two most recent denoising timesteps for each non-keyframe. For frame $f \in \mathcal{F}_n$ at layer $\ell$, the cache $\mathcal{C}_\ell^{(f)}$ maintains:
\begin{equation}
\mathcal{C}_\ell^{(f)} = \left\{
\begin{aligned}
&\mathbf{h}_{\text{before}}^{(\ell, f)} \in \mathbb{R}^{N \times D}, \quad &&t_{\text{before}}, \\
&\mathbf{h}_{\text{after}}^{(\ell, f)} \in \mathbb{R}^{N \times D}, \quad &&t_{\text{after}},
\end{aligned}
\right\}
\end{equation}
where $\mathbf{h}_{\text{before}}^{(\ell, f)}$ and $\mathbf{h}_{\text{after}}^{(\ell, f)}$ are the cached attention outputs at the two most recent denoising steps $t_{\text{before}} > t_{\text{after}}$ (noting that diffusion timesteps decrease during denoising). Crucially, we cache the \textbf{output hidden states} (post-attention) rather than input latents, as they already incorporate contextual information from all other frames.

At the end of the warmup phase (step $t_{\text{w}}$), we initialize the cache for all non-keyframes in all layers:
\begin{equation}
\begin{split}
    \mathbf{h}_{\text{before}}^{(\ell, f)} = \mathbf{h}_{\text{after}}^{(\ell, f)} = \text{Attention}^{(\ell)}\left(\mathbf{Q}_{t_{\text{warmup}}}^{(f)}, \mathbf{K}_{t_{\text{warmup}}}, \mathbf{V}_{t_{\text{warmup}}}\right), \\t_{\text{before}} = t_{\text{after}} = t_{\text{w}}.
\end{split}
\end{equation}


At a skipped timestep $t_{\text{skip}}$ where $t_{\text{after}} < t_{\text{skip}} < t_{\text{before}}$, we reconstruct the frame representation via latent trajectories projection:
\begin{equation}
\mathbf{h}_{t_{\text{skip}}}^{(\ell, f)} = (1 - \alpha) \cdot \mathbf{h}_{\text{before}}^{(\ell, f)} + \alpha \cdot \mathbf{h}_{\text{after}}^{(\ell, f)},
\end{equation}
where the interpolation weight $\alpha \in [0, 1]$ is computed based on the relative temporal position:
\begin{equation}
\alpha = \frac{t_{\text{before}} - t_{\text{skip}}}{t_{\text{before}} - t_{\text{after}} + \epsilon}, \quad \epsilon = 10^{-8}.
\end{equation}
This formulation ensures that $\alpha \to 0$ as $t_{\text{skip}} \to t_{\text{before}}$ (favoring the earlier cached state) and $\alpha \to 1$ as $t_{\text{skip}} \to t_{\text{after}}$ (favoring the later cached state), aligning with the decreasing-timestep denoising trajectory.

\paragraph{Attention computation at skipped steps:} While non-keyframes $f \in \mathcal{F}_n$ do not compute query outputs, they must still provide key-value pairs for keyframe attention. We directly use the latent projected hidden states:
\begin{equation}
\mathbf{K}_{t_{\text{skip}}}^{(f)} = \mathbf{V}_{t_{\text{skip}}}^{(f)} = \mathbf{h}_{t_{\text{skip}}}^{(\ell, f)}, \quad f \in \mathcal{F}_n.
\end{equation}
Keyframes $f \in \mathcal{F}_k$ perform full attention over all frames:
\begin{equation}
\begin{split}
    \mathbf{h}_{t_{\text{skip}}}^{(\ell, f)} = \text{Attention}^{(\ell)} ( \mathbf{Q}_{t_{\text{skip}}}^{(f)}, \left[\mathbf{K}_{t_{\text{skip}}}^{(1)}, \ldots, \mathbf{K}_{t_{\text{skip}}}^{(F)}\right], \\ \left[\mathbf{V}_{t_{\text{skip}}}^{(1)}, \ldots, \mathbf{V}_{t_{\text{skip}}}^{(F)}\right]), \quad f \in \mathcal{F}_k.
\end{split}
\end{equation}

\paragraph{Cache Update at Denoising Steps:} When non-keyframes perform denoising at "Rhythmic Point" $t_{\text{denoise}}$, we update the rolling cache via a shift-and-store strategy:

\begin{equation}
\begin{aligned}
\mathbf{h}_{\text{before}}^{(\ell, f)} &\leftarrow \mathbf{h}_{\text{after}}^{(\ell, f)}, \\
t_{\text{before}} &\leftarrow t_{\text{after}}, \\
\mathbf{h}_{\text{after}}^{(\ell, f)} &\leftarrow \text{Attn}^{(\ell)}\bigl(\mathbf{Q}_{t_{\text{denoise}}}^{(f)}, \mathbf{K}_{t_{\text{denoise}}}, \mathbf{V}_{t_{\text{denoise}}}\bigr), \\
t_{\text{after}} &\leftarrow t_{\text{denoise}}.
\end{aligned}
\end{equation}

This rolling update ensures that the cache always retains the two most recent denoising states, enabling accurate projection for the next $(n-1)$ skipped steps.

\subsubsection{Cross-Layer Consistency Guarantees.}

A subtle but critical challenge in multi-layer architectures is ensuring \textbf{temporal consistency} across layers. Inconsistent cache states can cause error propagation, as layer $\ell+1$ depends on the output of layer $\ell$. We enforce consistency through three mechanisms:

\begin{itemize}
\item \textbf{Shared global step counter}: A class-level variable is incremented only in layer~0 and shared across all layer instances, ensuring all layers agree on the current timestep.
\item \textbf{Synchronized cache updates}: All layers update their caches simultaneously at denoising steps, preventing temporal misalignment.
\item \textbf{Shared keyframe indices}: The set $\mathcal{F}_k$ is determined once at the end of warmup (layer~0) and frozen thereafter, ensuring all layers use the same denoising schedule.
\end{itemize}

Formally, let $\tau(\ell)$ denote the effective timestep used by layer $\ell$. Consistency requires $\tau(\ell) = \tau(\ell') = t_{\text{current}}$ for all $\ell, \ell'$, which our design guarantees.

\section{Detailed Implementation of RhymeFlow}
\label{sec:implementation}

To provide a clear procedural understanding of the proposed framework, we present the algorithmic details of \textit{RhymeFlow}. The process is divided into two primary stages: (1) Content-aware sequential keyframe selection, and (2) Asynchronous denoising flow scheduling.

\subsection{Sequential Keyframe Selection}

As discussed in Section 3.1 of the main paper, traditional uniform frame selection fails to capture non-linear semantic transitions, while computing similarity on noisy latents $\mathbf{z}_T$ is unreliable due to noise interference. We first perform a single-step denoising proxy to estimate the clean latents $\hat{\mathbf{z}}_0$ as a robust basis for selection. To prevent keyframes from clustering too densely or being too sparse due to a fixed similarity threshold, we propose a Dynamic Threshold Adjustment mechanism, as detailed in Algorithm \ref{alg:keyframe_selection}. We define an expected average interval $L_{avg} = N/M$. If the temporal gap between the current and previous keyframe significantly exceeds $L_{avg}$ (controlled by $\delta_{up}$), the threshold $\tau$ is decreased by $\Delta \tau$ to encourage selection; conversely, if the gap is too small (controlled by $\delta_{down}$), $\tau$ is increased to suppress redundant keyframes. This ensures a content-adaptive yet well-distributed keyframe set.

\begin{algorithm}[t]
    \caption{Dynamic Sequential Keyframe Selection}
    \label{alg:keyframe_selection}
    \begin{algorithmic}[1]
        \Require Initial noisy latents $\{\mathbf{z}_{T}^{(i)}\}_{i=1}^{N}$, Initial threshold $\tau$, Max keyframe budget $M$, Threshold step $\Delta \tau$, Interval tolerance $\{\delta_{up}, \delta_{down}\}$.
        \Ensure Keyframe set $\mathcal{K}$, Non-keyframe set $\mathcal{N}$.
        \State \textit{// Step 1: Predict clean latent proxies}
        \State $\{\hat{\mathbf{z}}_0^{(i)}\}_{i=1}^{N} \gets \text{Denoise}(\{\mathbf{z}_{T}^{(i)}\}, T \to 0)$ \Comment{Single-step approximation}
        
        \State \textit{// Step 2: Adaptive sequential selection}
        \State $\mathcal{K} \gets \{1\}$, $idx_{last} \gets 1$, $L_{avg} \gets N/M$ 
        \For{$i = 2$ \textbf{to} $N$}
            \If{$|\mathcal{K}| < M$}
                \State $S \gets \text{CosineSimilarity}(\hat{\mathbf{z}}_0^{(i)}, \hat{\mathbf{z}}_0^{(idx_{last})})$
                \If{$S < \tau$}
                    \State $\Delta L \gets i - idx_{last}$ \Comment{Calculate actual interval}
                    \State \textit{// Dynamic threshold adjustment}
                    \If{$\Delta L \ge L_{avg} + \delta_{up}$} 
                        \State $\tau \gets \tau - \Delta \tau$ \Comment{Decrease threshold to relax selection}
                    \ElseIf{$\Delta L \le L_{avg} - \delta_{down}$}
                        \State $\tau \gets \tau + \Delta \tau$ \Comment{Increase threshold to tighten selection}
                    \EndIf
                    \State $\mathcal{K} \gets \mathcal{K} \cup \{i\}$
                    \State $idx_{last} \gets i$
                \EndIf
            \EndIf
        \EndFor
        \State $\mathcal{N} \gets \{1, \dots, N\} \setminus \mathcal{K}$
        \State \Return $\mathcal{K}$, $\mathcal{N}$
    \end{algorithmic}
\end{algorithm}

\subsection{The Overall RhymeFlow Pipeline}

The core of \textit{RhymeFlow} lies in its heterogeneous scheduling. While keyframes maintain a standard step-by-step denoising trajectory to preserve structural integrity, non-keyframes follow an accelerated path by skipping redundant steps. Crucially, during skipped steps, we employ \textit{Latent Trajectory Projection} to synthesize the missing temporal context, ensuring that keyframe updates remain globally coherent. This process is summarized in Algorithm \ref{alg:rhymeflow}.

\begin{algorithm}[t]
    \caption{RhymeFlow: Asynchronous Denoising Flow Scheduling}
    \label{alg:rhymeflow}
    \begin{algorithmic}[1]
        \Require Noisy latents $\{\mathbf{z}_{T}^{(i)}\}$, Timesteps $T$, Warm-up steps $T_w$, Mid-point $T_{mid}$, Progressive strides $\{n_{small}, n_{large}\}$.
        \Ensure Fully denoised latents $\{\mathbf{z}_0^{(i)}\}$.
        
        \State $\mathcal{K}, \mathcal{N} \gets \text{DynamicKeyframeSelection}(\dots)$ \Comment{Via Algorithm \ref{alg:keyframe_selection}}
        
        \For{$t = T$ \textbf{down to} $1$}
            \If{$t > T - T_w$} \Comment{\textbf{Phase I: Synchronous Warm-up}}
                \State $\{\mathbf{z}_{t-1}^{(i)}\}_{i=1}^{N_f} \gets \text{FullDenoiseStep}(\{\mathbf{z}_{t}^{(i)}\}_{i=1}^{N_f})$ \Comment{Full 3D Attention}
            \Else \Comment{\textbf{Phase II: Asynchronous Scheduling}}
                \State \textit{// Progressive Stride Assignment (Eq. 2)}
                \State $n_{skip} \gets (t > T_{mid}) ? n_{small} : n_{large}$ 
                
                \State \textit{// Identify active frames to be updated}
                \State $\mathcal{I}_{active} \gets \mathcal{K} \cup \{j \in \mathcal{N} \mid (T - T_w - t) \equiv 0 \pmod{n_{skip}}\}$
                
                \For{each frame $i \in \mathcal{I}_{active}$}
                    \State \textit{// Latent Trajectory Projection (Sec 3.2)}
                    \State $\mathcal{C} \gets \{\mathbf{z}_t^{(k)}\}_{k \in \mathcal{K}} \cup \{\hat{\mathbf{z}}_t^{(j)}\}_{j \in \mathcal{N} \setminus \mathcal{I}_{active}}$ \Comment{Anchor + Projected states}
                    
                    \State $n_i \gets (i \in \mathcal{K}) ? 1 : n_{skip}$ \Comment{Step size for key/non-keyframe}
                    \State $\mathbf{z}_{t-n_i}^{(i)} \gets \text{Denoise}(\mathbf{z}_t^{(i)} \mid \mathcal{C})$ \Comment{Heterogeneous update}
                \EndFor
                \State \textit{// Skipped non-keyframes $\{\mathbf{z}^{(j)}\}$ remain at state $t$ until next rhythmic point.}
            \EndIf
        \EndFor
        \State \Return $\{\mathbf{z}_0^{(i)}\}_{i=1}^{N_f}$
    \end{algorithmic}
\end{algorithm}

\section{More Discussions}

\subsection{Analysis of Failure Cases}

\begin{figure}[htbp]
  \centering
  \includegraphics[width=1.0\linewidth]{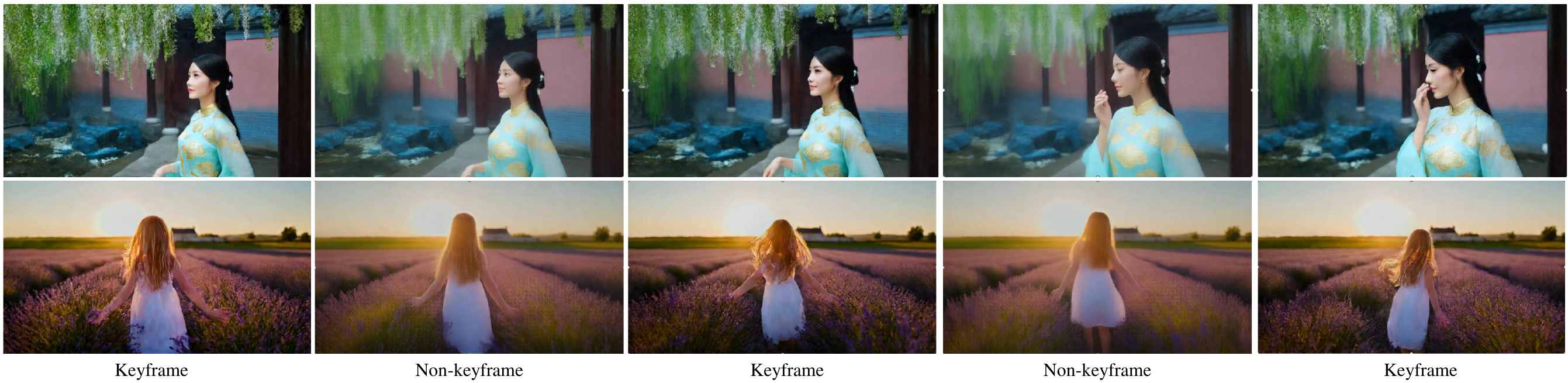}
   \caption{Visual failure cases under extreme acceleration ($n_{skip}=7$). Aggressive skipping leads to shimmering artifacts and loss of high-frequency textures in non-keyframes, highlighting the necessity of our progressive scheduling balance.}
   \label{fig:re2}
\end{figure}

To explore the upper bound of our acceleration framework, we conducted stress tests with an ultra-aggressive skipping stride ($n_{skip}=7$). As illustrated in Figure~\ref{fig:re2}, such extreme skipping triggers noticeable temporal aliasing and "shimmering" effects, particularly in non-keyframe regions. This degradation occurs because the linear approximation provided by our \textit{Latent Trajectory Projection} assumes a locally smooth ODE path; at very large strides, this assumption fails to capture high-frequency motion non-linearities, causing asynchronous frames to drift from the anchor keyframes. These failure cases validate that our default conservative scheduling ($n_{small}=2, n_{large}=3$) is essential for maintaining artifact-free temporal coherence.

\subsection{Future Work}

Several promising avenues exist for extending our work. A primary direction is the exploration of \textit{learned asynchronous scheduling}. While our current progressive strategy is heuristic-based, future research could employ a small policy network or uncertainty-based metric to dynamically determine which frames should be skipped and at what intervals, enabling a truly data-dependent acceleration.

Additionally, we envision the integration of \textit{RhymeFlow} with post-training quantization (PTQ) and low-rank adaptations (LoRA). Since our method is strictly training-free and operates at the scheduling level, it remains orthogonal to model-level compression techniques. Combining these two paradigms could potentially unlock $3\times$ to $5\times$ speedups without compromising the generative capabilities of large-scale Video DiTs. Finally, investigating more sophisticated interpolation techniques beyond linear projection—such as utilizing optical flow priors—could further suppress artifacts at even larger skip strides.

\bibliographystyle{splncs04}
\bibliography{main}